\documentclass{article} 
\usepackage{iclr2025_conference,times}

\usepackage{amsmath,amsfonts,bm}

\def\eqref#1{equation~\ref{#1}}

\def\1{\bm{1}}

\DeclareMathAlphabet{\mathsfit}{\encodingdefault}{\sfdefault}{m}{sl}
\SetMathAlphabet{\mathsfit}{bold}{\encodingdefault}{\sfdefault}{bx}{n}

\usepackage{hyperref}
\usepackage{url}

\title{Neural Catalog: Scaling Species Recognition with Catalog of Life–Augmented Generation}

\iclrfinalcopy

\author{Faizan Farooq Khan$^1$,   \, Jun Chen$^1$, \, 
Youssef Mohamed1$^1$, \, Chun-Mei Feng$^2$ \textbf{Mohamed Elhoseiny}$^1$ \\
$^1$King Abdullah University of Science and Technology\\
$^2$ IHPC, A*STAR \\
\texttt{{\{faizan.khan, jun.chen,youssef.mohamed,mohamed.elhoseiny\}}@kaust.edu.sa}\\
\texttt{\{fengcm.ai\}@gmail.com}
\footnotesize
}

\newcommand{\pok}{{Pokémon}}

\usepackage{booktabs}
\usepackage{cleveref}

\usepackage{amsmath}
\usepackage{pgfplots}

\usepackage{adjustbox}
\usepackage{wrapfig}
\usepackage{pgfplotstable}
\usepackage{array}
\usepackage{nicefrac}
\usepackage{subcaption}
\usepackage{wrapfig}
\usetikzlibrary{calc}
\usepackage[table]{xcolor}
\usepackage{multirow}
\usepackage{pifont}

\usepackage[noend]{algpseudocode}
\usepackage{algorithm}

\definecolor{blond}{rgb}{0.98, 0.94, 0.75}

\begin{document}

\maketitle

\begin{abstract}
Open-vocabulary species recognition is a major challenge in computer vision, particularly in ornithology, where new taxa are continually discovered. While benchmarks like CUB-200-2011 and Birdsnap have advanced fine-grained recognition under closed vocabularies, they fall short of real-world conditions. We show that current systems suffer a performance drop of over 30\% in realistic open-vocabulary settings with thousands of candidate species, largely due to an increased number of visually similar and semantically ambiguous distractors. To address this, we propose Visual Re-ranking Retrieval-Augmented Generation (VR-RAG), a novel framework that links structured encyclopedic knowledge with recognition. We distill Wikipedia articles for 11,202 bird species into concise, discriminative summaries and retrieve candidates from these summaries. Unlike prior text-only approaches, VR-RAG incorporates visual information during retrieval, ensuring final predictions are both textually relevant and visually consistent with the query image. Extensive experiments across five bird classification benchmarks and two additional domains show that VR-RAG improves the average performance of the state-of-the-art Qwen2.5-VL model by 18.0\%.

\end{abstract}
\section{Introduction}
\label{sec:intro}

The American Museum of Natural History estimates a total of 18,043 bird species~\cite{barrowclough2016many}. Meanwhile, the CUB-200-2011~\cite{cub} dataset, a popular dataset for bird species classification, contains only 200 classes. The drastic gap questions the real-world applicability of benchmarks based on small, fixed taxonomies. Such concern goes beyond bird classification to other categories and objects. Compounding this challenge, more species are discovered annually\footnote{popularmechanics.com/science/animals/a30501204/new-bird-species-discovered}. This problem naturally becomes more challenging as new species continue to be discovered each year. The Catalogue of Life-2025 saw 48,766 newly accepted species names added~\cite{cof}. Accordingly, there is a pressing need to develop AI classification systems that can flexibly handle these challenges while providing competitive performance on existing classes.

The problem of recognizing species from an ever-increasing set is recognized as open-vocabulary recognition~\cite{wu2023open}. Open-vocabulary recognition is characterized by having unseen classes during test time that the model has no idea about, and the label space itself may evolve. Learning about all the bird species is practically infeasible due to limited annotations. Therefore, we need methods that can utilize additional knowledge to recognize new species classes at test time.

A straightforward way of classifying new unseen classes in the open-vocabulary paradigm is using class descriptions. The concept of visually classifying objects from textual descriptions, along with the associated contrastive learning algorithm, can be traced back to~\cite{writeaclass}. And with recent and scalable \textbf {V}ision \textbf{L}anguage \textbf{M}odels (\textbf{VLM}s) like CLIP~\cite{clip}, this task has seen rapid progress. While CLIP-like models can handle a large number and variety of classes, they underperform on open-vocabulary setups. Previous works~\cite{clip, openclip, siglip} have evaluated the capabilities of CLIP-like models to recognize unseen classes in scenarios where the label space is restricted to species present in the dataset. In Table~\ref{tab:mot}, we show that when expanding the vocabulary to $11,202$ species, the performance of CLIP drops by an average of 31.5\%(35.5\% across five benchmarks).

\begin{wraptable}[10]{b}{0.5\textwidth} 
\vspace{-4mm}
  \centering
  \caption{Average accuracy drop when moving from a closed-world setting (species present in dataset) to an open-world setting with all 11,202 species.}
  \centering{
\footnotesize
\scalebox{0.75}{
\begin{tabular}{l|cc|cc|c}
Model & \multicolumn{2}{c|}{CUB} & \multicolumn{2}{c|}{Birdsnap}  \\
\midrule
 & Bounded & Open & Bounded & Open &  Average  Drop\\

\midrule
CLIP & 62.9 & 24.6 & 51.3 & 26.5 & 31.5 \\ 
OpenCLIP & 74.3 & 35.2 & 61.4 & 33.7 & 33.4 \\
SigLIP & 77.2 & 49.9 & 64.6 & 43.6 & 24.1 \\
\midrule
\end{tabular}
}
}
\label{tab:mot}
\end{wraptable}

While using detailed descriptions sees a similar performance pattern, it is a more promising direction for two key reasons. First, from a practical standpoint, descriptions are highly accessible: non-experts can visually describe a newly discovered species, enabling identification before a formal taxonomic name is even assigned. Second, this approach is future-proof:  as new species are discovered and as we get stronger vision foundation models, these models can perform better with no or minimal changes by exploiting the rich detail in descriptions, overcoming the knowledge cutoffs that limit today's systems. Yet, current VLMs like CLIP~\cite{clip} are limited: trained on static image-caption pairs, they cannot easily integrate new knowledge and struggle with subtle morphological reasoning. We believe their shortcomings arise from a reliance on overly simplistic prompts (e.g., ``a photo of a [species]'') and an inability to reason about the subtle morphological differences that rich descriptions can provide.

To overcome these limitations, we turn to Multimodal Large Language Models (MLLMs). Instead of relying on static, pre-trained knowledge, we integrate a dynamic external knowledge base through Retrieval-Augmented Generation (RAG), allowing the model to access detailed, up-to-date information on demand. However, naively applying RAG to fine-grained species is ineffective due to three critical problems: (1) the sheer scale of avian biodiversity (over 11,000 species) makes indexing and retrieving from comprehensive sources like Wikipedia infeasible at inference time; (2) generic taxonomic descriptions often lack the visual details needed for discriminative recognition; and (3) the retrieved content can exceed the context length of MLLMs~\cite{Yin_2024}.


We address these challenges with Visual Re-ranking RAG (VR-RAG), a novel framework that uses structured encyclopedic knowledge with vision-language reasoning. First, we curate a comprehensive benchmark comprising Wikipedia articles for $11,202$ bird species, distilling each into concise, visually salient summaries using GPT-4o~\cite{gpt4o} to eliminate non-discriminative text (e.g., conservation status), and extracting the visual information available and filtering them using GPT-4o.  Second, we design a RAG pipeline that retrieves candidate species based on multimodal similarity. This is followed by a visual re-ranking stage that refines results using Dino-v2~\cite{dinov2} similarities. Finally, we augment MLLMs with the retrieved refined summaries, enabling them to reason about species identity by correlating key visual traits (e.g., "wing color", "leg color", etc.).

The principles underlying our framework are domain-agnostic and extend beyond avian classification. We demonstrate this by evaluating our method on marine species recognition on the FishNet~\cite{fishnet} dataset, and on our Pokémon dataset. We curate the \pok~dataset, which includes over $1,000$ Pokémon species descriptions. Experiments across five bird benchmarks, FishNet~\cite{fishnet}, and our Pokémon dataset demonstrate the superiority of our framework. We outperform CLIP by 35.9\% on MRR@1 in the retrieval task and by 39.4\% in species recognition accuracy while eliminating the need for dataset-specific retraining. In addition, we improve by 18.0\% over the best MLLM model. Our contributions can be summarized as follows: 
\begin{itemize}
    \item We demonstrate that a visual re-ranking module within Retrieval Augmented Generation can substantially mitigate noisy retrieval in large-scale, fine-grained recognition. Our proposed VR-RAG improves retrieval precision (mRR@10) by 27.4\% over the strongest baseline.
    \item We introduce an open-vocabulary benchmark for bird species recognition that spans $11,202$ species. Unlike prior datasets limited to a few hundred categories, our benchmark leverages Wikipedia-derived multimodal knowledge and GPT-4-refined summaries, enabling evaluation at realistic biodiversity scales.
    \item We show that integrating VR-RAG yields a substantial boost to state-of-the-art MLLMs for open-world species recognition, improving the average accuracy of the top-performing model by 18.0\% across five challenging bird benchmarks.
    \item We further validate the cross-domain versatility of VR-RAG by applying it to FishNet and our curated \pok~dataset, where it consistently outperforms baselines. This demonstrates that our framework generalizes beyond birds to other fine-grained recognition domains.
\end{itemize}


\section{Related Work}
\label{sec:related}

\paragraph{Open-Vocabulary Recognition.} 
Early work on open-vocabulary recognition~\cite{ogovc} introduced joint image–word embeddings. With the rise of multimodal pre-training in NLP (e.g., BERT~\cite{bert}), vision–language models like CLIP~\cite{clip} soon emerged, later extended to detection~\cite{vild}, segmentation~\cite{lseg}, and classification~\cite{ocls,anytime}. Stronger variants like OpenCLIP~\cite{openclip} and SigLIP~\cite{siglip} achieved impressive zero-shot classification but still struggle in open-world species recognition due to limited taxonomic knowledge and dataset bias. We systematically evaluate these models and show that our approach surpasses existing baselines in open-vocabulary bird recognition.



\paragraph{Species Recognition.} 
Species recognition is a long-standing challenge in fine-grained image classification, with birds as a central benchmark. Datasets like CUB-200-2011~\cite{cub}, Birdsnap~\cite{birdsnap}, and iNaturalist~\cite{inat} have advanced the field, but evaluations largely remain closed-set, restricted to predefined species~\cite{clip, writeaclass, ferjad1}. In reality, recognition is open-world, where many species are unseen during training. In this work, we move to an open-vocabulary setting across five benchmarks~\cite{cub, birdsnap, indian_birds, inat, nabirds}, performing recognition over the full set of species from Wikipedia.


\paragraph{MultiModal Large Language Models(MLLM).} 
MLLMs have advanced significantly in understanding and reasoning across modalities~\cite{gpt4o, gpt4v, gemini, deepseek, qwen25, mixtralexperts, llama}. Scaling in data and models, combined with pre-training, supervised fine-tuning, and reinforcement learning from human feedback~\cite{rlhf}, has enabled strong emergent reasoning abilities~\cite{qwen25}. However, MLLMs still degrade in long contexts~\cite{Yin_2024}, motivating retrieval-based approaches that extend context capacity while mitigating hallucinations.


\paragraph{Retrieval Augmented Generation (RAG).} 
RAG enhances large models by integrating external knowledge via retrieval systems~\cite{clip, siglip, openclip, noisyclip, faiss, colbert}. The original framework by~\cite{lewis} jointly optimized a retriever and generator, with later work refining fusion mechanisms~\cite{izacard} and extending RAG to long-form reasoning, multi-hop QA, and hallucination mitigation~\cite{hal1, asai2023selfrag}. Vision-centric extensions include MuRAG~\cite{murag}, which uses an image-text memory bank, MIRAGE~\cite{mirage} employs a CLIP-based retriever, and REVEAL~\cite{reveal} leverages multimodal graphs for reasoning. We propose VR-RAG, a two-stage framework that fuses multiple vision encoders and applies a re-ranker to refine top-$k$ candidates, achieving superior performance over existing methods.


\section{Dataset and Benchmark}
\label{sec:data}

Our benchmark comprises 11,202 species, each paired with a corresponding Wikipedia article, a concise summary describing the species, and representative images referred to as anchor images.

\paragraph{Data Collection.}
We begin by compiling a comprehensive list of all bird species available on Wikipedia~\footnote{https://en.wikipedia.org/wiki/List\_of\_birds\_by\_common\_name}. For each species, we retrieve all information from its respective Wikipedia page, both textual and visual. Species without dedicated pages are excluded from our dataset. This process results in a curated set of $11,202$ bird species, each accompanied by its Wikipedia-sourced description and visual information, forming a rich knowledge base for our open-vocabulary recognition task. 

\paragraph{Summary Generation and Visual Refinement.}
Wikipedia articles often contain extraneous information, such as name origins and historical context, which are useless for visual discrimination. This irrelevant content can degrade the performance by increasing the context length without adding meaningful visual cues. To address this, we employ GPT-4o~\cite{gpt4o} to refine the Wikipedia articles by generating concise summaries focused on attributes relevant to distinguishing bird species visually. We do so by prompting GPT-4o~\cite{gpt4o} with the species Wikipedia article and prompting it with: \textit{Summarize the following information about the bird species ``species name'' into a concise paragraph. Focus on the key physical characteristics that distinguish this species from other similar bird species. Highlight features like size, beak shape, plumage color patterns, wing shape, and any other unique traits. The summary should be useful for someone trying to identify the bird species from a photograph.} This reduces the average word count per species from $552$ to $127$.  

The visual content extracted from Wikipedia articles is often heterogeneous, including not only photographs of the species but also non-target materials such as habitat maps, conservation park logos, and scientific illustrations. To filter the authentic photographs, each image was fed to GPT-4o and evaluated with the direct prompt:``Is there a bird present in the Image? Reply only with Yes or No". Only images that received a `Yes' response were retained. The retained images serve as representative ``anchor images" for each species. After filtering, each species had an average of three anchor images associated with it.

\paragraph{Pokémon dataset.} We built our Pokémon dataset through a multi-stage pipeline. We began with the canonical list of all Pokémon species from \href{https://pokemondb.net/tools/text-list}{Pokémondb text list}. For each species, we used GPT-4o to generate a concise, descriptive summary highlighting key visual characteristics from Wikipedia in a similar fashion as done for birds. Image collection was performed by scraping results from Google Search and multiple Kaggle datasets based on \pok. However, we noticed the scraped images often contained multiple \pok s or were of the wrong species. To ensure overall image quality and consistency, all scraped images were manually verified against a representative image for each species. The image was selected from \href{https://pokemondb.net/pokedex/all}{Pokémondb Pokedex page}. All the inaccurate samples were discarded. This curation process yielded a final test set of 11,216 images spanning 248 species. 

\paragraph{Description Quality.} 
We evaluate the description quality by performing human evaluation in two ways:
1) The users are shown four images of the corresponding species along with the description, and they are asked to rate the description on a 1-5 scale, with 1 indicating "not helpful at all" and 5 indicating "very helpful" for identifying the species. This was done for 10\% of the bird species descriptions and all 248 descriptions from the \pok dataset. Each description was rated by three human annotators. Bird descriptions received a mean score of 4.3, a median of 4.0, and a mode of 5.0, while \pok descriptions received a mean score of 4.5, a median of 5.0, and a mode of 5.0. For consistency, we calculated inter-annotator reliability using Quadratic Weighted Agreement (QWA). Across both tasks, we achieved a QWA of 90.6\% for bird descriptions and 92.2\% for \pok~descriptions. The complete breakdown of the calculation of the reliability scores can be found in~\cref{app:human}.
2) We conduct a matching test, where the users are shown a description and five images, and they are asked to choose the image that is most matching. To ensure the task is fine-grained, we conducted this test on the 200 species from the cub dataset. We ensure the correct image is always present in the five images, and the remaining four are chosen as the closest to the description in the clip space. On average, the users spent almost one minute on each description and achieved 94\% accuracy. For the remaining 6\%, 4\% cases were where the users could not converge to one single option, and the remaining 2\% were wrong. We discuss and showcase qualitative samples for all the cases with provided explanation in~\cref{app:human}.

\begin{figure*}[h]
    \centering
    \includegraphics[width=\linewidth]{figs/fig1.pdf}
        \caption{\textbf{The VR-RAG pipeline}. Left: Data extraction, both textual and visual, from Wikipedia articles. Middle: Similarity calculation with the query image using an ensemble of multi-modal encoders. Right: Re-ranker module using anchor images for final similarity score calculation.}
    \label{fig:1}
\end{figure*}  

    
\section{Preliminaries}
\label{sec:prelim}

This section introduces the problem setting and provides the necessary background knowledge.

\subsection{Task Formulation}  
We focus on \textit{open-vocabulary species recognition}, which aims to determine the species of a given image. Unlike traditional classification tasks, where the label space is restricted to species present in the training database~\cite{survey}, our task considers an open and ever-expanding textual label space, making the problem inherently cross-modal.

\subsection{Vision-Language Models (VLMs)}
Vision-Language Models (VLMs) like CLIP~\cite{clip} map text $y$ and images $x$ into a shared representation space using a text encoder $f(\cdot)$ and vision encoder $g(\cdot)$.
VLMs are trained on large-scale image-text datasets, such as WIT-400M~\cite{wit} and LAION~\cite{laion}, using contrastive learning, where text representations $f(y)$ and image representations $g(x)$ of corresponding caption-image pairs are aligned. While effective for many vision tasks~\cite{vlmsurvey}, VLMs alone are insufficient for our task, as they lack the reasoning capabilities to differentiate between similar species~\cite{clipun1, clipun2}. Therefore, additional reasoning mechanisms are needed to improve recognition performance. 

\subsection{Multi-Modal Large Language Models (MLLMs)} 
MLLMs such as~\cite{qwen25, internvl,Yin_2024} are trained on massive multimodal datasets, enabling them to handle various vision and language tasks. These models possess reasoning capabilities and can integrate information from both textual and visual inputs. However, they are still prone to hallucinations when dealing with large contexts~\cite{jin2025longcontext}. To mitigate this, RAG has been proposed to provide up-to-date information as context.. 
However, the performance gains from RAG are highly dependent on the design of an effective multimodal retrieval mechanism.

\section{Method}
\label{sec:method}

Given a query image $q$ and a set of $\mathcal{N}$ species summaries and associated anchor images $a_i$, our approach consists of a retrieval module followed by a reranker module and then an MLLM module. First, we retrieve the top $r$ species summary chunks using our retrieval module. These chunks are then re-ranked using a visual re-ranker module to obtain the top $k$ chunks. The chunks are then mapped to their respective species summaries, which are subsequently provided to the MLLM along with the query image $q$ to determine the species present in the query image.

\subsection{Retrieval}  
To adhere to the context length limitations of our models, we first segment each of the $\mathcal{N}$ species summaries into smaller text chunks. For each species, its set of $K$ anchor images $\{a_i\}_{i=1}^K$ are processed by a visual encoder $g(\cdot)$, and their embeddings are averaged to create a single visual representation. Concurrently, each associated text chunk, $t$, is processed by a text encoder $f(\cdot)$ to produce a text embedding. Each chunk is represented by its multi-modal representation, $m$ by:
\begin{equation}
    m =  \frac{ f(t) + \frac{1}{K} \sum_{i=1}^{K} g(a_i) }{2}
\end{equation}
We then compute the similarity between a given query image $q$ and each multi-modal representation $m$. This multi-modal similarity $s^m$, is calculated as the dot product between the query image's embedding $g(q)$ and the multi-modal embedding $m$, $s^{m} = g(q)^\top m$. All embeddings are $l_2$-normalised.

Ensemble methods have been shown to improve performance across various machine learning tasks~\cite{ensemble}. We observe a similar trend with VLMs, leading us to employ an ensemble for improved retrieval. Specifically, we combine the similarity scores from CLIP, OpenCLIP, and SigLIP by averaging their similarity scores to obtain a final ranking. This ensemble approach effectively leverages the unique strengths of each VLM, resulting in better retrieval performance than any single model could achieve on its own. This is shown in the central part of~\cref{fig:1}

\subsection{Reranker}  
After retrieving the top $r$ chunks, we refine the selection further by re-ranking them to obtain the top $k$ chunks. This re-ranking step utilizes both textual relevance and visual similarity. For each species represented in the top $r$ chunks, we sample the anchor images. We then calculate the intra-modal similarity between the anchors $a_i$ and query $q$. For our vision encoder $h$, we use Dino-v2~\cite{dinov2}, which has demonstrated strong performance in visual understanding-related tasks~\cite{eyes, camb, longvu}. The final similarity score for each chunk is obtained by a combination of intra-modal and the previously calculated multi-modal similarity weighted by $\lambda$, as defined in~\cref{eqn:sims}. The top $k$ chunks, after re-ranking, are selected for the next stage. A visual approach to our re-ranker module is presented in~\cref{fig:rerank}.

\begin{equation}
\begin{split}
    s^i &= h(q)^\top \left(\frac{1}{K} \sum_{i=1}^{K} h(a_i) \right) \\
    s &= \lambda s^m + (1 - \lambda)s^i
\end{split}
    \label{eqn:sims}
\end{equation}

\subsection{MLLM}  

Instead of passing only the top-$k$ retrieved chunks, we input the full summaries of the $k$ species associated with the top-$k$ ranked chunks to the MLLM, along with the query image $q$. This is feasible because our summaries are concise and discriminative, unlike lengthy Wikipedia articles that often are filled with non-discriminative details. As a result, even when $k$ increases (e.g., 5, 10, or 15), the MLLM can process all candidate summaries within its context window and make a more informed decision about the species in $q$ without being overwhelmed by irrelevant information. This is ablated in~\cref{tab:sum_ablation} where we show that providing our refined summary works best by improving 4.0\% over raw Wikipedia articles and 12.4\% over using only retrieved chunks as context.

\section{Experiments and Results}
\label{sec:res}

Below, we outline our experimental setup, detailing the datasets used, evaluation metrics, and baseline methods. We then present our results, accompanied by an ablation study, to demonstrate the effectiveness of each component.

\subsection{Experimental Setup}

\paragraph{Datasets.} We evaluate our approach on five bird classification datasets. \textbf{CUB-200-2011}~\cite{cub} consists of 11,788 images of 200 bird species from North and South America, with 5,794 images used for testing. \textbf{Birdsnap}~\cite{birdsnap} is a larger dataset of North American birds, containing 49,829 images spanning 500 species, with 2,443 images allocated for testing. \textbf{Indian Birds}~\cite{indian_birds} includes 25 species with a test set of 7,499 images. \textbf{iNaturalist}~\cite{inat} is a large-scale biodiversity dataset covering thousands of plant and animal species, designed for real-world fine-grained classification challenges. We extract all images of the ‘Aves’ class from the validation set, resulting in 13,230 images spanning 1,323 bird species. \textbf{NaBirds}~\cite{nabirds} is a fine-grained bird classification dataset focused on North American species. It comprises 48,562 annotated images across 555 visual categories, which correspond to 400 unique bird species, with $24,633$ images for testing. At test time, all 11,202 collected species summaries are ranked for each image, aligning the evaluation with an open-vocabulary setting.

We also evaluate VR-RAG on two additional datasets. The first contains marine species from the \textbf{FishNet}~\cite{fishnet} dataset, where we follow the original split strategy for species classification. To align with our open-vocabulary setting, each query image is compared against 17,393 candidate descriptions sourced from FishNet. For the \textbf{\pok}~dataset, we evaluate on the full set of 248 species using 1,024 descriptions.

\paragraph{Metrics.} For retrieval evaluation, we use mean reciprocal rank (mRR), which is the average inverse rank of the first retrieved ground truth image. We report mRR@1, mRR@5, and mRR@10 to evaluate the ranking of the first correct among the top retrieved chunks. For recognition, we report the percentage of correctly classified examples to measure performance.

\paragraph{Baselines.} In our experiment, we evaluate several VLMs for retrieval and both VLMs and MLLMs for recognition tasks. For the retrieval problem, we use CLIP~\cite{clip}, OpenCLIP~\cite{openclip}, and SigLIP~\cite{siglip}, all of which employ the ViT-L14 as the backbone. We use BioClip~\cite{bioclip}, which employs the ViT-B16 backbone. 

For the recognition task, we evaluate the same VLMs used for retrieval. We follow the standard protocol of calculating similarity and selecting the highest-rated chunk. The chunk is mapped to the species it belongs to evaluate CLIP-like models. Although these models are trained to match object categories with images, our setting involves working with summaries. Therefore, we compute the similarity between the query image and summary chunks for a fair evaluation. 

We assess multiple MLLMs, including Qwen2-VL-7B Instruct from the Qwen2-VL~\cite{qwen2} suite, Qwen2.5-VL-7B Instruct from Qwen2.5-VL~\cite{qwen25}, Gemma-3n-e4b-it~\cite{gemma3}, Mini-CPM-V-2.9~\cite{minicpm} and InternVL3-8B from Intern-VL~\cite{internvl3}. We assess these models in two settings, one by directly asking them the species name given a query image, and the second by providing summaries for the top $k$ species retrieved via VR-RAG as context. We select $r$ as 100, $k$ as 5, and $\lambda$ as 0.7 for our experiments. 

\begin{table*}[b!]
\centering
\setlength\tabcolsep{12pt}
\caption{Retrieval Results: We compare VR-RAG with other cross-modal(text-to-image) retrieval methods across the five benchmarks. VR-RAG consistently outperforms the baseline models on mRR@1, mRR@5, and mRR@10 metrics. The best results are highlighted in bold text. }
\scalebox{0.60}{
\begin{tabular}{lccccccccc}
\toprule
 & \multicolumn{3}{c}{\textbf{Birdsnap}}&  \multicolumn{3}{c}{\textbf{CUB}} & \multicolumn{3}{c}{\textbf{iNaturalist Birds}} \\
 & mRR@1 & mRR@5 & mRR@10   & mRR@1 & mRR@5 & mRR@10 & mRR@1 & mRR@5 & mRR@10  \\
\cmidrule(r){1-1} 
\cmidrule(lr){2-4} \cmidrule(lr){5-7} \cmidrule(lr){8-10}
BioCLIP & 19.4 & 26.5 & 27.7 & 24.3 & 31.2 & 32.4 & 13.5 & 19.2 & 20.2 \\
CLIP & 17.2 & 25.1 & 26.5 & 16.1 & 24.1 & 25.6 & 7.7 & 12.3 & 13.4 \\
OpenCLIP & 15.7 & 24.8 & 26.3 & 15.2 & 23.9 & 25.4 & 8.5 & 13.7 & 14.9 \\
SigLIP & 18.8 & 27.9 & 29.3 & 20.1 & 29.6 & 31.1 & 11.3 & 17.2 & 18.3 \\
\rowcolor{blond}
\textbf{VR-RAG(ours)} & \textbf{48.9} & \textbf{52.6} & \textbf{53.8} & \textbf{58.0} & \textbf{62.3} & \textbf{63.1} & \textbf{34.8} & \textbf{38.7} & \textbf{39.8} \\
\midrule
 & \multicolumn{3}{c}{\textbf{Indian Birds}} & \multicolumn{3}{c}{ \textbf{NABirds}} &
 \multicolumn{3}{c}{\textbf{Average}} \\
  & mRR@1 & mRR@5 & mRR@10 &  mRR@1 & mRR@5 & mRR@10 & mRR@1 & mRR@5 & mRR@10\\
\cmidrule(r){1-1} 
\cmidrule(lr){2-4} \cmidrule(lr){5-7} \cmidrule(lr){8-10}

BioCLIP & 23.0 & 29.2 & 30.0 & 21.8 & 29.1 & 30.3 & 20.4 & 27.0 & 28.1 \\
CLIP & 17.6 & 25.9 & 27.2 & 18.2 & 26.4 & 27.9 & 15.4 & 22.8 & 24.1 \\
OpenCLIP & 10.2 & 16.8 & 18.3 & 15.5 & 23.9 & 25.6 & 13.0 & 20.6 & 22.1 \\
SigLIP & 21.9 & 30.4 & 31.9 & 20.5 & 29.6 & 31.1 & 18.5 & 26.9 & 28.3 \\
\rowcolor{blond}
\textbf{VR-RAG(ours)}  & \textbf{68.1} & \textbf{72.9} & \textbf{73.6} & \textbf{52.0} & \textbf{56.6} & \textbf{57.6} & \textbf{52.3} & \textbf{56.6} & \textbf{57.6}\\
\bottomrule
\end{tabular}
}
\label{tab:retrival}
\vspace{-0.3cm}
\end{table*}

\subsection{Results}

\paragraph{Retrieval.} We present the retrieval results in the ~\cref{tab:retrival}. Across all five benchmarks, VR-RAG outperforms the baseline methods. It achieves the highest scores for all levels (1, 5, 10) on the mRR metric. On average, VR-RAG improves upon the best baseline (SigLIP~\cite{siglip}) by 32.8\% on MRR@1, 27.8\% on MRR@5, and 27.4\% on MRR@10. These improvements highlight the strength of our approach, which combines three vision-language encoders and a visual re-ranker.

\begin{wraptable}[16]{t}{0.5\textwidth} 
\vspace{-5mm}
\caption{Classification Accuracy: We evaluate open-source MLLMs directly and integrate them with our VR-RAG pipeline. We also show results for various VLMs. Direct RAG uses CLIP~\cite{clip} as the RAG module.}
\vspace{-3mm}
\centering
\scalebox{0.65}{
\footnotesize
\begin{tabular}{l|rrrrrlr}
& \rotatebox{90}{Birdsnap} & \rotatebox{90}{CUB} & \rotatebox{90}{iNat.}     & \rotatebox{90}{Ind. Birds}   & \rotatebox{90}{NABirds} & \rotatebox{90}{Average} \\
\midrule
BioCLIP~\cite{bioclip} & 19.4 & 24.3 & 13.5 & 23.0 & 21.8 & 20.4 \\
CLIP~\cite{clip} & 17.2 & 16.1 & 7.7 & 17.6 & 18.2 & 15.4 \\
OpenCLIP~\cite{openclip} & 15.7 & 15.2 & 8.5 & 10.2 & 15.5 & 13.0 \\
SigLIP~\cite{siglip} & 18.8 & 20.1 & 11.3 & 21.9 & 20.5 & 18.5 \\
InternVL-3~\cite{internvl3} & 12.7 & 14.9 & 4.0 & 0.5 & 14.4 & 9.3 \\
MiniCPM-V-2.6~\cite{minicpm} & 9.4 & 8.0 & 2.8 & 1.8 & 10.8 & 6.6 \\
Qwen2VL~\cite{qwen2} & 34.3 & 37.2 & 14.7 & 24.0 & 42.7 & 30.6 \\
Gemma-3n~\cite{qwen2} & 17.2 & 19.8 & 8.5 & 15.6 & 22.8 & 16.8 \\
Qwen2.5VL~\cite{qwen25} & 39.3 & 44.7 & 17.9 & 33.1 & 49.2 & 36.8 \\
Qwen 2.5VL + direct RAG & 29.5 & 29.2 & 13.2 & 29.5 & 31.7 & 26.6 \\
\rowcolor{blond}
\midrule
Qwen 2VL + VR-RAG (ours) & 49.8 & 57.3 & 34.1 & 70.2 & 55.5 & 53.4 \\
\rowcolor{blond}
Qwen 2.5VL + VR-RAG (ours) & \textbf{51.9} & \textbf{60.3} & \textbf{35.5} & \textbf{72.2} & \textbf{56.3} & \textbf{55.2} \\
\midrule
\end{tabular}
}
\label{tab:baseline}
\end{wraptable}

\paragraph{Recognition.} In~\cref{tab:baseline}, we present the recognition results across all five benchmarks. A consistent performance improvement is observed across MLLMs when augmented with VR-RAG. In contrast, direct RAG negatively impacts QWEN2.5-VL, which we attribute to the inferior retrieval capabilities of CLIP compared to our VR-RAG module. Specifically, VR-RAG enhances QWEN2-VL by 22.3\%, and QWEN2.5-VL by 18.0\%.
We also evaluated GPT-4o on the iNat-Birds dataset, as it has the most extensive species coverage among the five datasets. In a direct zero-shot setting, GPT-4o achieved 23.9\% accuracy, outperforming all open-source models tested. When augmented with our VR-RAg framework, its performance increased substantially to 36.0\%. Notably, this is only slightly higher than the performance of the augmented Qwen2.5-VL, highlighting the strength of open-source models that are significantly smaller than GPT-4o.
Notably, VLMs such as CLIP~\cite{clip} perform poorly due to their inability to effectively reason over detailed summaries~\cite{clipun1, clipun2} and their limited context window~\cite{longclip}. VR-RAG, on the other hand, provides a flexible and easily integrable framework for species recognition, making it particularly advantageous in dynamic scenarios where species may be newly discovered or become extinct. This eliminates the need for full model retraining, ensuring adaptability in open-world settings. Results for the remaining MLLMs with VR-RAG are presented in~\cref{app:addvrrag}.

\begin{wraptable}[6]{t}{0.4\textwidth} 
\vspace{-8mm}
\caption{Classification accuracy for the FishNet and \pok~datasets.}
\vspace{-2mm}
\centering
\scalebox{0.70}{
\footnotesize
\begin{tabular}{l|rrl}
& FishNet & \pok \\
\midrule
SigLIP~\cite{siglip} & 3.1 & 75.5 \\
Qwen2.5VL~\cite{qwen25} & 5.0 & 29.4 \\
Qwen 2.5VL + direct RAG & 4.5 & 60.5 \\
\rowcolor{blond}
\midrule
\rowcolor{blond}
Qwen 2.5VL + VR-RAG (ours) & \textbf{18.7} & \textbf{86.3} \\
\midrule
\end{tabular}
}
\label{tab:domain}
\end{wraptable}

\subsection{Additional Domains}
We present results for FishNet and the \pok~dataset in~\cref{tab:domain}, highlighting the best-performing MLLM and VLM. Both VLMs and MLLMs perform poorly on FishNet, likely due to the lack of marine species knowledge and the extremely fine-grained nature of the task, which involves a large vocabulary of over $17,000$ species. While these results are unsurprising, our approach still improves upon the best MLLM by 12.9\%. For the \pok~dataset, SigLIP outperforms QWEN2.5-VL by a huge gap(46.1\%). But, when augmented by VR-RAG, it improves by 56.2\%, clearly showcasing that VR-RAG is highly effective across domains. Detailed evaluations on both datasets are provided in~\cref{app:fishpok}.

\subsection{Ablation}

\begin{table*}[b!]
\centering
\setlength\tabcolsep{10pt}
\caption{Ablation study: We evaluate each of VR-RAG module's impact for the MRR@1, MRR@5, and MRR@10 on CUB-200-2011~\cite{cub}, Birdsnap~\cite{birdsnap}, and Indian Birds~\cite{indian_birds}. The best results are highlighted in bold text for each column.}

\scalebox{0.6}{
\begin{tabular}{ccccccccccc}
\toprule
 \multirow{2}{*}{\textbf{CLIP}}  & \multirow{2}{*}{ \textbf{OpenCLIP}} &
 \multirow{2}{*}{\textbf{SigLIP}} &
 \multirow{2}{*}{\textbf{Multimodal}} &
 \multirow{2}{*}{\textbf{Re-Ranker}}\multirow{2}{*} &\multicolumn{3}{c}{\textbf{Birdsnap~\cite{birdsnap}}} &\multicolumn{3}{c}{\textbf{CUB~\cite{cub}}} \\
 &&&&& MRR@1 & MRR@5 & MRR@10 &MRR@1 & MRR@5 & MRR@10 \\

\cmidrule(r){1-1} 
\cmidrule(lr){2-2} \cmidrule(lr){3-3} \cmidrule(lr){4-4} \cmidrule(lr){5-5} 
\cmidrule(lr){6-8} \cmidrule(lr){9-11}
 

{\color{green} \checkmark} & {\color{red} \ding{55}} & {\color{red} \ding{55}} & {\color{red} \ding{55}} & {\color{red} \ding{55}} & 17.2 & 25.1 & 26.5 & 16.1 & 24.1 & 25.6\\



{\color{green} \checkmark} & {\color{green} \checkmark} & {\color{red} \ding{55}}& {\color{red} \ding{55}} & {\color{red} \ding{55}} & 25.5 & 35.5 & 36.9 & 25.5 & 36.0 & 37.4 \\

{\color{green} \checkmark} & {\color{green} \checkmark} & {\color{green} \checkmark} & {\color{red} \ding{55}} & {\color{red} \ding{55}} & 28.8 & 39.2 & 40.5 & 30.0 & 40.9 & 42.4 \\

{\color{green} \checkmark} & {\color{green} \checkmark} & {\color{green} \checkmark} & {\color{green} \checkmark} & {\color{red} \ding{55}} & 40.6 & 47.3 & 48.5 & 40.9 & 42.4 & 48.6\\

{\color{green} \checkmark} & {\color{green} \checkmark} & {\color{green} \checkmark} & {\color{green} \checkmark} & {\color{green} \checkmark} & {\textbf{48.9}} & {\textbf{52.6}} & {\textbf{53.8}} &
{\textbf{58.0}} &{\textbf{62.3}} &{\textbf{63.1}}\\

\bottomrule
\end{tabular}
}
\vspace{-0.1cm}
\label{tab:retr_ablate}
\end{table*}

\paragraph{Retrieval Components.} The ablation study in~\cref{tab:retr_ablate} illustrates the contribution of each component in the VR-RAG pipeline. CLIP alone performs poorly across all benchmarks, but incremental additions of OpenCLIP and SigLIP lead to consistent improvements. The best encoder performance is achieved when all three are combined. Incorporating multi-modal embeddings, comprising textual summaries and visual anchors, further boosts performance across all datasets. Finally, employing Dino-V2 to rerank the retrieved chunks results in significant gains across all mRR levels. Overall, the study confirms that every component of VR-RAG is essential to its effectiveness. An ablation of MLLM performance when attached to each of the retrieval pipelines from~\cref{tab:retr_ablate} is presented in~\cref{app:abl_mllm_ret}, and ablations for VLMs and re-rankers are presented in~\cref{app:vlm_rerank_abl}.

\begin{wrapfigure}[14]{t}{0.4\textwidth} 
\vspace{-8mm}
    \centering
    \includegraphics[width=\linewidth]{figs/fig2.pdf}
        \caption{Impact of varying top-k candidates($5$ to $15$) fed to Qwen2.5-VL.}
    \label{fig:2}
    \vspace{-5mm}
\end{wrapfigure} 

\paragraph{Top-$k$ Selection.} In~\cref{fig:2}, we analyze the impact of increasing the number of top-k species candidates from $k=5$ to $k=15$ on performance. The data reveal a consistent inverse relationship between context length and accuracy, suggesting that the models struggle to identify the relevant information within a longer candidate list. 

For Qwen2.5-VL, average accuracy steadily declines from 54.8\% at $k=5$ to 54.3\% at $k=10$, and further to 52.8\% at $k=15$. This drop of two percentage points confirms that the expanded context with more options hinders the model's reasoning. This decline in performance indicates that by providing more candidate species, the models appear to be overwhelmed by the increased number of options, making it more challenging to focus on the subtle, fine-grained details needed for accurate species classification. This highlights a key limitation in these vision-language models for this specific task: their inability to effectively filter out redundant or confusing information when presented with a large set of candidates. It also emphasizes the importance of providing a concise, high-quality context rather than a large volume of raw data. 
Despite the drop, it is noteworthy that even at $k=15$, the performance(52.8\%) of Qwen2.5-VL is superior to providing raw Wikipedia articles(51.2\%).

\begin{wrapfigure}[7]{t}{0.3\textwidth} 
\vspace{-9mm}
    \centering
    \includegraphics[width=\linewidth]{figs/lambda_ablation.png}
    \vspace{-8mm}
        \caption{Impact of varying $\lambda$.}
    \label{fig:lamda}
    \vspace{-5mm}
\end{wrapfigure}

\paragraph{Varying $\lambda$.} The final re-ranked similarity score is a weighted average of the multi-modal and the intra-modal similarity from DinoV2, with the balance controlled by the parameter $\lambda$. As shown in~\cref{fig:lamda}, the optimal performance is achieved at $\lambda=0.7$. This indicates that slightly favoring the multi-modal similarity yields the best results, while relying too heavily on an individual similarity (as $\lambda$ approaches 0 or 1) leads to suboptimal performance.

\begin{wraptable}[8]{t}{0.3\textwidth} 
\vspace{-4mm}
\caption{Studying the impact of different types of context fed to MLLM.}
\vspace{-2mm}
\centering
\scalebox{0.5}{
\footnotesize
\begin{tabular}{l|rrrrrlr}
& \rotatebox{90}{CUB} & \rotatebox{90}{Birdsnap} & \rotatebox{90}{iNat}     & \rotatebox{90}{Ind. Birds}   & \rotatebox{90}{NABirds} & \rotatebox{90}{Average} \\
\midrule

Chunks & 41.7 & 46.9 & 26.0 & 52.1 & 47.2 & 42.8 \\
Raw Wikipedia & 47.7 & 56.4 & 33.6 & 65.8 & 52.7 & 51.2 \\
\rowcolor{blond}
Ours & \textbf{51.9} & \textbf{60.3} & \textbf{35.5} & \textbf{72.2} & \textbf{56.3} & \textbf{55.2} \\
\midrule
\end{tabular}
\label{tab:vlm_abl}
}

\label{tab:sum_ablation}
\end{wraptable}

\paragraph{Ablating summary refinement pipeline.} 
As shown in~\cref{tab:sum_ablation}, QWEN2.5-VL performs poorly when given only top-ranked chunks, as they lack sufficient detail to distinguish the query image. Using the full Wikipedia article improves accuracy but adds large amounts of irrelevant text, increasing noise and processing cost. In contrast, our refined summaries strike a balance by retaining only discriminative details, enabling better reasoning and yielding the strongest performance.


\section{Conclusion}
\label{sec:conclusion}

In this work, we focus on the task of open-vocabulary species recognition and demonstrate the limitations of current Vision-Language Models (VLMs) in this setting. To address these challenges, we propose VR-RAG, a multimodal re-ranking retrieval-augmented generation (RAG) framework that significantly improves retrieval performance over a large pool of species summaries.  

Additionally, we collect and refine Wikipedia summaries for $11,202$ bird species, distilling them into short, discriminative descriptions. We curate a \pok~dataset with descriptions for over $1,000$ \pok s. Experimental results show that augmenting VR-RAG with MLLMs leads to superior performance across all five avian, one marine, and one \pok~benchmark.

We believe our work contributes to species recognition in open-world settings. Future work will focus on developing models capable of reasoning whether the provided species summary lacks critical information for species identification. Additionally, these models should autonomously retrieve supplementary details from web sources when needed.

\bibliography{iclr2025_conference}

\begin{thebibliography}{61}
\providecommand{\natexlab}[1]{#1}
\providecommand{\url}[1]{\texttt{#1}}
\expandafter\ifx\csname urlstyle\endcsname\relax
  \providecommand{\doi}[1]{doi: #1}\else
  \providecommand{\doi}{doi: \begingroup \urlstyle{rm}\Url}\fi

\bibitem[Abbasi et~al.(2025)Abbasi, Rohban, and Baghshah]{clipun2}
Reza Abbasi, Mohammad~Hossein Rohban, and Mahdieh~Soleymani Baghshah.
\newblock Deciphering the role of representation disentanglement: Investigating compositional generalization in clip models.
\newblock In Ale{\v{s}} Leonardis, Elisa Ricci, Stefan Roth, Olga Russakovsky, Torsten Sattler, and G{\"u}l Varol (eds.), \emph{Computer Vision -- ECCV 2024}, pp.\  35--50, Cham, 2025. Springer Nature Switzerland.
\newblock ISBN 978-3-031-73024-5.

\bibitem[AI(2024)]{gemini}
Google AI.
\newblock Gemini: Google's multimodal ai model.
\newblock \emph{Google AI Research}, 2024.
\newblock \url{https://fireflies.ai/blog/gemini-vs-gpt-4}.

\bibitem[Asai et~al.(2023)Asai, Wu, Wang, Sil, and Hajishirzi]{asai2023selfrag}
Akari Asai, Zeqiu Wu, Yizhong Wang, Avirup Sil, and Hannaneh Hajishirzi.
\newblock {Self-RAG}: Learning to retrieve, generate, and critique through self-reflection.
\newblock \emph{arXiv preprint arXiv:2310.11511}, 2023.
\newblock URL \url{https://arxiv.org/abs/2310.11511}.

\bibitem[Barrowclough et~al.(2016)Barrowclough, Cracraft, Klicka, and Zink]{barrowclough2016many}
George~F Barrowclough, Joel Cracraft, John Klicka, and Robert~M Zink.
\newblock How many kinds of birds are there and why does it matter?
\newblock \emph{PLoS One}, 11\penalty0 (11):\penalty0 e0166307, 2016.

\bibitem[Berg et~al.(2014)Berg, Liu, Lee, Alexander, Jacobs, and Belhumeur]{birdsnap}
Thomas Berg, Jiongxin Liu, Seung~Woo Lee, Michelle~L. Alexander, David~W. Jacobs, and Peter~N. Belhumeur.
\newblock Birdsnap: Large-scale fine-grained visual categorization of birds.
\newblock In \emph{Proc. Conf. Computer Vision and Pattern Recognition (CVPR)}, June 2014.

\bibitem[Chen et~al.(2022)Chen, Hu, Chen, Verga, and Cohen]{murag}
Wenhu Chen, Hexiang Hu, Xi~Chen, Pat Verga, and William~W. Cohen.
\newblock Murag: Multimodal retrieval-augmented generator for open question answering over images and text, 2022.
\newblock URL \url{https://arxiv.org/abs/2210.02928}.

\bibitem[Chen et~al.(2024)Chen, Wu, Wang, Su, Chen, Xing, Zhong, Zhang, Zhu, Lu, et~al.]{internvl}
Zhe Chen, Jiannan Wu, Wenhai Wang, Weijie Su, Guo Chen, Sen Xing, Muyan Zhong, Qinglong Zhang, Xizhou Zhu, Lewei Lu, et~al.
\newblock Internvl: Scaling up vision foundation models and aligning for generic visual-linguistic tasks.
\newblock In \emph{Proceedings of the IEEE/CVF Conference on Computer Vision and Pattern Recognition}, pp.\  24185--24198, 2024.

\bibitem[Cohen(1968)]{Cohen1968WeightedKN}
Jacob Cohen.
\newblock Weighted kappa: Nominal scale agreement provision for scaled disagreement or partial credit.
\newblock \emph{Psychological Bulletin}, 70:\penalty0 213--220, 1968.
\newblock URL \url{https://api.semanticscholar.org/CorpusID:29694079}.

\bibitem[{COL FPC}(2024)]{cof}
{COL FPC}.
\newblock Catalogue of life.
\newblock URL: \url{https://www.catalogueoflife.org}, 2024.
\newblock DOI: 10.48580/dgt98.

\bibitem[Dao et~al.(2023)Dao, Huynh, Zhao, Phung, and Cai]{ocls}
Son~D. Dao, Dat Huynh, He~Zhao, Dinh Phung, and Jianfei Cai.
\newblock Open-vocabulary multi-label image classification with pretrained vision-language model.
\newblock In \emph{2023 IEEE International Conference on Multimedia and Expo (ICME)}, pp.\  2135--2140, 2023.
\newblock \doi{10.1109/ICME55011.2023.00365}.

\bibitem[DeepSeek-AI et~al.(2025)DeepSeek-AI, Guo, Yang, Zhang, Song, Zhang, Xu, Zhu, Ma, Wang, Bi, Zhang, Yu, Wu, Wu, Gou, Shao, Li, Gao, Liu, Xue, Wang, Wu, Feng, Lu, Zhao, Deng, Zhang, Ruan, Dai, Chen, Ji, Li, Lin, Dai, Luo, Hao, Chen, Li, Zhang, Bao, Xu, Wang, Ding, Xin, Gao, Qu, Li, Guo, Li, Wang, Chen, Yuan, Qiu, Li, Cai, Ni, Liang, Chen, Dong, Hu, Gao, Guan, Huang, Yu, Wang, Zhang, Zhao, Wang, Zhang, Xu, Xia, Zhang, Zhang, Tang, Li, Wang, Li, Tian, Huang, Zhang, Wang, Chen, Du, Ge, Zhang, Pan, Wang, Chen, Jin, Chen, Lu, Zhou, Chen, Ye, Wang, Yu, Zhou, Pan, Li, Zhou, Wu, Ye, Yun, Pei, Sun, Wang, Zeng, Zhao, Liu, Liang, Gao, Yu, Zhang, Xiao, An, Liu, Wang, Chen, Nie, Cheng, Liu, Xie, Liu, Yang, Li, Su, Lin, Li, Jin, Shen, Chen, Sun, Wang, Song, Zhou, Wang, Shan, Li, Wang, Wei, Zhang, Xu, Li, Zhao, Sun, Wang, Yu, Zhang, Shi, Xiong, He, Piao, Wang, Tan, Ma, Liu, Guo, Ou, Wang, Gong, Zou, He, Xiong, Luo, You, Liu, Zhou, Zhu, Xu, Huang, Li, Zheng, Zhu, Ma, Tang, Zha, Yan, Ren, Ren, Sha, Fu, Xu, Xie, Zhang,
  Hao, Ma, Yan, Wu, Gu, Zhu, Liu, Li, Xie, Song, Pan, Huang, Xu, Zhang, and Zhang]{deepseek}
DeepSeek-AI, Daya Guo, Dejian Yang, Haowei Zhang, Junxiao Song, Ruoyu Zhang, Runxin Xu, Qihao Zhu, Shirong Ma, Peiyi Wang, Xiao Bi, Xiaokang Zhang, Xingkai Yu, Yu~Wu, Z.~F. Wu, Zhibin Gou, Zhihong Shao, Zhuoshu Li, Ziyi Gao, Aixin Liu, Bing Xue, Bingxuan Wang, Bochao Wu, Bei Feng, Chengda Lu, Chenggang Zhao, Chengqi Deng, Chenyu Zhang, Chong Ruan, Damai Dai, Deli Chen, Dongjie Ji, Erhang Li, Fangyun Lin, Fucong Dai, Fuli Luo, Guangbo Hao, Guanting Chen, Guowei Li, H.~Zhang, Han Bao, Hanwei Xu, Haocheng Wang, Honghui Ding, Huajian Xin, Huazuo Gao, Hui Qu, Hui Li, Jianzhong Guo, Jiashi Li, Jiawei Wang, Jingchang Chen, Jingyang Yuan, Junjie Qiu, Junlong Li, J.~L. Cai, Jiaqi Ni, Jian Liang, Jin Chen, Kai Dong, Kai Hu, Kaige Gao, Kang Guan, Kexin Huang, Kuai Yu, Lean Wang, Lecong Zhang, Liang Zhao, Litong Wang, Liyue Zhang, Lei Xu, Leyi Xia, Mingchuan Zhang, Minghua Zhang, Minghui Tang, Meng Li, Miaojun Wang, Mingming Li, Ning Tian, Panpan Huang, Peng Zhang, Qiancheng Wang, Qinyu Chen, Qiushi Du, Ruiqi Ge, Ruisong
  Zhang, Ruizhe Pan, Runji Wang, R.~J. Chen, R.~L. Jin, Ruyi Chen, Shanghao Lu, Shangyan Zhou, Shanhuang Chen, Shengfeng Ye, Shiyu Wang, Shuiping Yu, Shunfeng Zhou, Shuting Pan, S.~S. Li, Shuang Zhou, Shaoqing Wu, Shengfeng Ye, Tao Yun, Tian Pei, Tianyu Sun, T.~Wang, Wangding Zeng, Wanjia Zhao, Wen Liu, Wenfeng Liang, Wenjun Gao, Wenqin Yu, Wentao Zhang, W.~L. Xiao, Wei An, Xiaodong Liu, Xiaohan Wang, Xiaokang Chen, Xiaotao Nie, Xin Cheng, Xin Liu, Xin Xie, Xingchao Liu, Xinyu Yang, Xinyuan Li, Xuecheng Su, Xuheng Lin, X.~Q. Li, Xiangyue Jin, Xiaojin Shen, Xiaosha Chen, Xiaowen Sun, Xiaoxiang Wang, Xinnan Song, Xinyi Zhou, Xianzu Wang, Xinxia Shan, Y.~K. Li, Y.~Q. Wang, Y.~X. Wei, Yang Zhang, Yanhong Xu, Yao Li, Yao Zhao, Yaofeng Sun, Yaohui Wang, Yi~Yu, Yichao Zhang, Yifan Shi, Yiliang Xiong, Ying He, Yishi Piao, Yisong Wang, Yixuan Tan, Yiyang Ma, Yiyuan Liu, Yongqiang Guo, Yuan Ou, Yuduan Wang, Yue Gong, Yuheng Zou, Yujia He, Yunfan Xiong, Yuxiang Luo, Yuxiang You, Yuxuan Liu, Yuyang Zhou, Y.~X. Zhu,
  Yanhong Xu, Yanping Huang, Yaohui Li, Yi~Zheng, Yuchen Zhu, Yunxian Ma, Ying Tang, Yukun Zha, Yuting Yan, Z.~Z. Ren, Zehui Ren, Zhangli Sha, Zhe Fu, Zhean Xu, Zhenda Xie, Zhengyan Zhang, Zhewen Hao, Zhicheng Ma, Zhigang Yan, Zhiyu Wu, Zihui Gu, Zijia Zhu, Zijun Liu, Zilin Li, Ziwei Xie, Ziyang Song, Zizheng Pan, Zhen Huang, Zhipeng Xu, Zhongyu Zhang, and Zhen Zhang.
\newblock Deepseek-r1: Incentivizing reasoning capability in llms via reinforcement learning, 2025.
\newblock URL \url{https://arxiv.org/abs/2501.12948}.

\bibitem[Devlin et~al.(2019)Devlin, Chang, Lee, and Toutanova]{bert}
Jacob Devlin, Ming-Wei Chang, Kenton Lee, and Kristina Toutanova.
\newblock Bert: Pre-training of deep bidirectional transformers for language understanding, 2019.
\newblock URL \url{https://arxiv.org/abs/1810.04805}.

\bibitem[Elhoseiny et~al.(2013)Elhoseiny, Saleh, and Elgammal]{writeaclass}
Mohamed Elhoseiny, Babak Saleh, and Ahmed Elgammal.
\newblock Write a classifier: Zero-shot learning using purely textual descriptions.
\newblock In \emph{2013 IEEE International Conference on Computer Vision}, pp.\  2584--2591, 2013.
\newblock \doi{10.1109/ICCV.2013.321}.

\bibitem[Gu et~al.(2022)Gu, Lin, Kuo, and Cui]{vild}
Xiuye Gu, Tsung-Yi Lin, Weicheng Kuo, and Yin Cui.
\newblock Open-vocabulary object detection via vision and language knowledge distillation.
\newblock In \emph{International Conference on Learning Representations}, 2022.
\newblock URL \url{https://openreview.net/forum?id=lL3lnMbR4WU}.

\bibitem[Hansen \& Salamon(1990)Hansen and Salamon]{ensemble}
L.K. Hansen and Peter Salamon.
\newblock Neural network ensembles.
\newblock \emph{Pattern Analysis and Machine Intelligence, IEEE Transactions on}, 12:\penalty0 993 -- 1001, 11 1990.
\newblock \doi{10.1109/34.58871}.

\bibitem[Hinchliff et~al.(2015)Hinchliff, Smith, Allman, Burleigh, Chaudhary, Coghill, Crandall, Deng, Drew, Gazis, Gude, Hibbett, Katz, Laughinghouse, McTavish, Midford, Owen, Ree, Rees, Soltis, Williams, and Cranston]{treeoflife}
C.~E. Hinchliff, S.~A. Smith, J.~F. Allman, J.~G. Burleigh, R.~Chaudhary, L.~M. Coghill, K.~A. Crandall, J.~Deng, B.~T. Drew, R.~Gazis, K.~Gude, D.~S. Hibbett, L.~A. Katz, H.~D.~4th Laughinghouse, E.~J. McTavish, P.~E. Midford, C.~L. Owen, R.~H. Ree, J.~A. Rees, D.~E. Soltis, T.~Williams, and K.~A. Cranston.
\newblock Synthesis of phylogeny and taxonomy into a comprehensive tree of life.
\newblock \emph{Proceedings of the National Academy of Sciences of the United States of America}, 112\penalty0 (41):\penalty0 12764--12769, Oct 13 2015.
\newblock \doi{10.1073/pnas.1423041112}.

\bibitem[Hu et~al.(2023)Hu, Iscen, Sun, Wang, Chang, Sun, Schmid, Ross, and Fathi]{reveal}
Ziniu Hu, Ahmet Iscen, Chen Sun, Zirui Wang, Kai-Wei Chang, Yizhou Sun, Cordelia Schmid, David~A. Ross, and Alireza Fathi.
\newblock Reveal: Retrieval-augmented visual-language pre-training with multi-source multimodal knowledge memory, 2023.
\newblock URL \url{https://arxiv.org/abs/2212.05221}.

\bibitem[Ilharco et~al.(2021)Ilharco, Wortsman, Wightman, Gordon, Carlini, Taori, Dave, Shankar, Namkoong, Miller, Hajishirzi, Farhadi, and Schmidt]{openclip}
Gabriel Ilharco, Mitchell Wortsman, Ross Wightman, Cade Gordon, Nicholas Carlini, Rohan Taori, Achal Dave, Vaishaal Shankar, Hongseok Namkoong, John Miller, Hannaneh Hajishirzi, Ali Farhadi, and Ludwig Schmidt.
\newblock Openclip, July 2021.
\newblock URL \url{https://doi.org/10.5281/zenodo.5143773}.
\newblock If you use this software, please cite it as below.

\bibitem[Izacard \& Grave(2021)Izacard and Grave]{izacard}
Gautier Izacard and Edouard Grave.
\newblock Leveraging passage retrieval with generative models for open domain question answering.
\newblock In Paola Merlo, Jorg Tiedemann, and Reut Tsarfaty (eds.), \emph{Proceedings of the 16th Conference of the European Chapter of the Association for Computational Linguistics: Main Volume}, pp.\  874--880, Online, April 2021. Association for Computational Linguistics.
\newblock \doi{10.18653/v1/2021.eacl-main.74}.
\newblock URL \url{https://aclanthology.org/2021.eacl-main.74/}.

\bibitem[Jia et~al.(2021)Jia, Yang, Xia, Chen, Parekh, Pham, Le, Sung, Li, and Duerig]{noisyclip}
Chao Jia, Yinfei Yang, Ye~Xia, Yi-Ting Chen, Zarana Parekh, Hieu Pham, Quoc~V. Le, Yunhsuan Sung, Zhen Li, and Tom Duerig.
\newblock Scaling up visual and vision-language representation learning with noisy text supervision, 2021.
\newblock URL \url{https://arxiv.org/abs/2102.05918}.

\bibitem[Jiang et~al.(2024)Jiang, Sablayrolles, Roux, Mensch, Savary, Bamford, Chaplot, de~las Casas, Hanna, Bressand, Lengyel, Bour, Lample, Lavaud, Saulnier, Lachaux, Stock, Subramanian, Yang, Antoniak, Scao, Gervet, Lavril, Wang, Lacroix, and Sayed]{mixtralexperts}
Albert~Q. Jiang, Alexandre Sablayrolles, Antoine Roux, Arthur Mensch, Blanche Savary, Chris Bamford, Devendra~Singh Chaplot, Diego de~las Casas, Emma~Bou Hanna, Florian Bressand, Gianna Lengyel, Guillaume Bour, Guillaume Lample, Lélio~Renard Lavaud, Lucile Saulnier, Marie-Anne Lachaux, Pierre Stock, Sandeep Subramanian, Sophia Yang, Szymon Antoniak, Teven~Le Scao, Théophile Gervet, Thibaut Lavril, Thomas Wang, Timothée Lacroix, and William~El Sayed.
\newblock Mixtral of experts, 2024.
\newblock URL \url{https://arxiv.org/abs/2401.04088}.

\bibitem[Jin et~al.(2025)Jin, Yoon, Han, and Arik]{jin2025longcontext}
Bowen Jin, Jinsung Yoon, Jiawei Han, and Sercan~O Arik.
\newblock Long-context {LLM}s meet {RAG}: Overcoming challenges for long inputs in {RAG}.
\newblock In \emph{The Thirteenth International Conference on Learning Representations}, 2025.
\newblock URL \url{https://openreview.net/forum?id=oU3tpaR8fm}.

\bibitem[Johnson et~al.(2021)Johnson, Douze, and Jégou]{faiss}
Jeff Johnson, Matthijs Douze, and Hervé Jégou.
\newblock Billion-scale similarity search with gpus.
\newblock \emph{IEEE Transactions on Big Data}, 7\penalty0 (3):\penalty0 535--547, 2021.
\newblock \doi{10.1109/TBDATA.2019.2921572}.

\bibitem[Khan et~al.(2023)Khan, Li, Temple, and Elhoseiny]{fishnet}
Faizan~Farooq Khan, Xiang Li, Andrew~J. Temple, and Mohamed Elhoseiny.
\newblock Fishnet: A large-scale dataset and benchmark for fish recognition, detection, and functional trait prediction.
\newblock In \emph{Proceedings of the IEEE/CVF International Conference on Computer Vision (ICCV)}, pp.\  20496--20506, October 2023.

\bibitem[Khattab \& Zaharia(2020)Khattab and Zaharia]{colbert}
Omar Khattab and Matei Zaharia.
\newblock Colbert: Efficient and effective passage search via contextualized late interaction over bert, 2020.
\newblock URL \url{https://arxiv.org/abs/2004.12832}.

\bibitem[Lewis et~al.(2020)Lewis, Perez, Piktus, Petroni, Karpukhin, Goyal, K\"{u}ttler, Lewis, Yih, Rockt\"{a}schel, Riedel, and Kiela]{lewis}
Patrick Lewis, Ethan Perez, Aleksandra Piktus, Fabio Petroni, Vladimir Karpukhin, Naman Goyal, Heinrich K\"{u}ttler, Mike Lewis, Wen-tau Yih, Tim Rockt\"{a}schel, Sebastian Riedel, and Douwe Kiela.
\newblock Retrieval-augmented generation for knowledge-intensive nlp tasks.
\newblock In H.~Larochelle, M.~Ranzato, R.~Hadsell, M.F. Balcan, and H.~Lin (eds.), \emph{Advances in Neural Information Processing Systems}, volume~33, pp.\  9459--9474. Curran Associates, Inc., 2020.
\newblock URL \url{https://proceedings.neurips.cc/paper_files/paper/2020/file/6b493230205f780e1bc26945df7481e5-Paper.pdf}.

\bibitem[Li et~al.(2022)Li, Weinberger, Belongie, Koltun, and Ranftl]{lseg}
Boyi Li, Kilian~Q Weinberger, Serge Belongie, Vladlen Koltun, and Rene Ranftl.
\newblock Language-driven semantic segmentation.
\newblock In \emph{International Conference on Learning Representations}, 2022.
\newblock URL \url{https://openreview.net/forum?id=RriDjddCLN}.

\bibitem[Mallen et~al.(2023)Mallen, Asai, Zhong, Das, Khashabi, and Hajishirzi]{hal1}
Alex Mallen, Akari Asai, Victor Zhong, Rajarshi Das, Daniel Khashabi, and Hannaneh Hajishirzi.
\newblock When not to trust language models: Investigating effectiveness of parametric and non-parametric memories.
\newblock pp.\  9802--9822, 01 2023.
\newblock \doi{10.18653/v1/2023.acl-long.546}.

\bibitem[Naeem et~al.(2023)Naeem, Ali~Khan, Xian, Afzal, Stricker, Van~Gool, and Tombari]{ferjad1}
Muhammad~Ferjad Naeem, Muhammad Gul~Zain Ali~Khan, Yongqin Xian, Muhammad~Zeshan Afzal, Didier Stricker, Luc Van~Gool, and Federico Tombari.
\newblock { I2MVFormer: Large Language Model Generated Multi-View Document Supervision for Zero-Shot Image Classification }.
\newblock In \emph{2023 IEEE/CVF Conference on Computer Vision and Pattern Recognition (CVPR)}, pp.\  15169--15179, Los Alamitos, CA, USA, June 2023. IEEE Computer Society.
\newblock \doi{10.1109/CVPR52729.2023.01456}.
\newblock URL \url{https://doi.ieeecomputersociety.org/10.1109/CVPR52729.2023.01456}.

\bibitem[OpenAI(2024{\natexlab{a}})]{gpt4o}
OpenAI.
\newblock Gpt-4o: Enhanced multimodal language model.
\newblock \emph{OpenAI Research}, 2024{\natexlab{a}}.
\newblock \url{https://openai.com/index/hello-gpt-4o/}.

\bibitem[OpenAI(2024{\natexlab{b}})]{gpt4v}
OpenAI.
\newblock Gpt-4v: Multimodal language model with vision capabilities.
\newblock \emph{OpenAI Research}, 2024{\natexlab{b}}.
\newblock \url{https://openai.com/index/gpt-4/}.

\bibitem[Oquab et~al.(2024)Oquab, Darcet, Moutakanni, Vo, Szafraniec, Khalidov, Fernandez, Haziza, Massa, El-Nouby, Assran, Ballas, Galuba, Howes, Huang, Li, Misra, Rabbat, Sharma, Synnaeve, Xu, Jegou, Mairal, Labatut, Joulin, and Bojanowski]{dinov2}
Maxime Oquab, Timothée Darcet, Théo Moutakanni, Huy Vo, Marc Szafraniec, Vasil Khalidov, Pierre Fernandez, Daniel Haziza, Francisco Massa, Alaaeldin El-Nouby, Mahmoud Assran, Nicolas Ballas, Wojciech Galuba, Russell Howes, Po-Yao Huang, Shang-Wen Li, Ishan Misra, Michael Rabbat, Vasu Sharma, Gabriel Synnaeve, Hu~Xu, Hervé Jegou, Julien Mairal, Patrick Labatut, Armand Joulin, and Piotr Bojanowski.
\newblock Dinov2: Learning robust visual features without supervision, 2024.
\newblock URL \url{https://arxiv.org/abs/2304.07193}.

\bibitem[Ouyang et~al.(2022)Ouyang, Wu, Jiang, Almeida, Wainwright, Mishkin, Zhang, Agarwal, Slama, Ray, Schulman, Hilton, Kelton, Miller, Simens, Askell, Welinder, Christiano, Leike, and Lowe]{rlhf}
Long Ouyang, Jeff Wu, Xu~Jiang, Diogo Almeida, Carroll~L. Wainwright, Pamela Mishkin, Chong Zhang, Sandhini Agarwal, Katarina Slama, Alex Ray, John Schulman, Jacob Hilton, Fraser Kelton, Luke Miller, Maddie Simens, Amanda Askell, Peter Welinder, Paul Christiano, Jan Leike, and Ryan Lowe.
\newblock Training language models to follow instructions with human feedback, 2022.
\newblock URL \url{https://arxiv.org/abs/2203.02155}.

\bibitem[Patel et~al.(2024)Patel, Kusumba, Cheng, Kim, Gokhale, Baral, and Yang]{clipun1}
Maitreya Patel, Abhiram Kusumba, Sheng Cheng, Changhoon Kim, Tejas Gokhale, Chitta Baral, and Yezhou Yang.
\newblock Tripletclip: Improving compositional reasoning of clip via synthetic vision-language negatives.
\newblock \emph{Advances in neural information processing systems}, 2024.

\bibitem[Qwen et~al.(2025)Qwen, :, Yang, Yang, Zhang, Hui, Zheng, Yu, Li, Liu, Huang, Wei, Lin, Yang, Tu, Zhang, Yang, Yang, Zhou, Lin, Dang, Lu, Bao, Yang, Yu, Li, Xue, Zhang, Zhu, Men, Lin, Li, Tang, Xia, Ren, Ren, Fan, Su, Zhang, Wan, Liu, Cui, Zhang, and Qiu]{qwen25}
Qwen, :, An~Yang, Baosong Yang, Beichen Zhang, Binyuan Hui, Bo~Zheng, Bowen Yu, Chengyuan Li, Dayiheng Liu, Fei Huang, Haoran Wei, Huan Lin, Jian Yang, Jianhong Tu, Jianwei Zhang, Jianxin Yang, Jiaxi Yang, Jingren Zhou, Junyang Lin, Kai Dang, Keming Lu, Keqin Bao, Kexin Yang, Le~Yu, Mei Li, Mingfeng Xue, Pei Zhang, Qin Zhu, Rui Men, Runji Lin, Tianhao Li, Tianyi Tang, Tingyu Xia, Xingzhang Ren, Xuancheng Ren, Yang Fan, Yang Su, Yichang Zhang, Yu~Wan, Yuqiong Liu, Zeyu Cui, Zhenru Zhang, and Zihan Qiu.
\newblock Qwen2.5 technical report, 2025.
\newblock URL \url{https://arxiv.org/abs/2412.15115}.

\bibitem[Radford et~al.(2021)Radford, Kim, Hallacy, Ramesh, Goh, Agarwal, Sastry, Askell, Mishkin, Clark, Krueger, and Sutskever]{clip}
Alec Radford, Jong~Wook Kim, Chris Hallacy, Aditya Ramesh, Gabriel Goh, Sandhini Agarwal, Girish Sastry, Amanda Askell, Pamela Mishkin, Jack Clark, Gretchen Krueger, and Ilya Sutskever.
\newblock Learning transferable visual models from natural language supervision, 2021.
\newblock URL \url{https://arxiv.org/abs/2103.00020}.

\bibitem[Rokde(2023)]{indian_birds}
Vaibhav Rokde.
\newblock Indian birds dataset, 2023.
\newblock URL \url{https://www.kaggle.com/datasets/ichhadhari/indian-birds}.
\newblock Accessed: 2025-01-04.

\bibitem[Schuhmann et~al.(2021)Schuhmann, Vencu, Beaumont, Kaczmarczyk, Mullis, Katta, Coombes, Jitsev, and Komatsuzaki]{wit}
Christoph Schuhmann, Richard Vencu, Romain Beaumont, Robert Kaczmarczyk, Clayton Mullis, Aarush Katta, Theo Coombes, Jenia Jitsev, and Aran Komatsuzaki.
\newblock Laion-400m: Open dataset of clip-filtered 400 million image-text pairs, 2021.
\newblock URL \url{https://arxiv.org/abs/2111.02114}.

\bibitem[Schuhmann et~al.(2022)Schuhmann, Beaumont, Vencu, Gordon, Wightman, Cherti, Coombes, Katta, Mullis, Wortsman, Schramowski, Kundurthy, Crowson, Schmidt, Kaczmarczyk, and Jitsev]{laion}
Christoph Schuhmann, Romain Beaumont, Richard Vencu, Cade Gordon, Ross Wightman, Mehdi Cherti, Theo Coombes, Aarush Katta, Clayton Mullis, Mitchell Wortsman, Patrick Schramowski, Srivatsa Kundurthy, Katherine Crowson, Ludwig Schmidt, Robert Kaczmarczyk, and Jenia Jitsev.
\newblock Laion-5b: An open large-scale dataset for training next generation image-text models, 2022.
\newblock URL \url{https://arxiv.org/abs/2210.08402}.

\bibitem[Shen et~al.(2024)Shen, Xiong, Zhao, Wu, Chen, Zhu, Liu, Xiao, Varadarajan, Bordes, Liu, Xu, Kim, Soran, Krishnamoorthi, Elhoseiny, and Chandra]{longvu}
Xiaoqian Shen, Yunyang Xiong, Changsheng Zhao, Lemeng Wu, Jun Chen, Chenchen Zhu, Zechun Liu, Fanyi Xiao, Balakrishnan Varadarajan, Florian Bordes, Zhuang Liu, Hu~Xu, Hyunwoo~J. Kim, Bilge Soran, Raghuraman Krishnamoorthi, Mohamed Elhoseiny, and Vikas Chandra.
\newblock Longvu: Spatiotemporal adaptive compression for long video-language understanding, 2024.
\newblock URL \url{https://arxiv.org/abs/2410.17434}.

\bibitem[Stevens et~al.(2024)Stevens, Wu, Thompson, Campolongo, Song, Carlyn, Dong, Dahdul, Stewart, Berger-Wolf, Chao, and Su]{bioclip}
Samuel Stevens, Jiaman Wu, Matthew~J Thompson, Elizabeth~G Campolongo, Chan~Hee Song, David~Edward Carlyn, Li~Dong, Wasila~M Dahdul, Charles Stewart, Tanya Berger-Wolf, Wei-Lun Chao, and Yu~Su.
\newblock {B}io{CLIP}: A vision foundation model for the tree of life.
\newblock In \emph{Proceedings of the IEEE/CVF Conference on Computer Vision and Pattern Recognition (CVPR)}, pp.\  19412--19424, 2024.

\bibitem[Team et~al.(2025)Team, Kamath, Ferret, Pathak, Vieillard, Merhej, Perrin, Matejovicova, Ramé, Rivière, Rouillard, Mesnard, Cideron, bastien Grill, Ramos, Yvinec, Casbon, Pot, Penchev, Liu, Visin, Kenealy, Beyer, Zhai, Tsitsulin, Busa-Fekete, Feng, Sachdeva, Coleman, Gao, Mustafa, Barr, Parisotto, Tian, Eyal, Cherry, Peter, Sinopalnikov, Bhupatiraju, Agarwal, Kazemi, Malkin, Kumar, Vilar, Brusilovsky, Luo, Steiner, Friesen, Sharma, Sharma, Gilady, Goedeckemeyer, Saade, Feng, Kolesnikov, Bendebury, Abdagic, Vadi, György, Pinto, Das, Bapna, Miech, Yang, Paterson, Shenoy, Chakrabarti, Piot, Wu, Shahriari, Petrini, Chen, Lan, Choquette-Choo, Carey, Brick, Deutsch, Eisenbud, Cattle, Cheng, Paparas, Sreepathihalli, Reid, Tran, Zelle, Noland, Huizenga, Kharitonov, Liu, Amirkhanyan, Cameron, Hashemi, Klimczak-Plucińska, Singh, Mehta, Lehri, Hazimeh, Ballantyne, Szpektor, Nardini, Pouget-Abadie, Chan, Stanton, Wieting, Lai, Orbay, Fernandez, Newlan, yeong Ji, Singh, Black, Yu, Hui, Vodrahalli, Greff, Qiu,
  Valentine, Coelho, Ritter, Hoffman, Watson, Chaturvedi, Moynihan, Ma, Babar, Noy, Byrd, Roy, Momchev, Chauhan, Sachdeva, Bunyan, Botarda, Caron, Rubenstein, Culliton, Schmid, Sessa, Xu, Stanczyk, Tafti, Shivanna, Wu, Pan, Rokni, Willoughby, Vallu, Mullins, Jerome, Smoot, Girgin, Iqbal, Reddy, Sheth, Põder, Bhatnagar, Panyam, Eiger, Zhang, Liu, Yacovone, Liechty, Kalra, Evci, Misra, Roseberry, Feinberg, Kolesnikov, Han, Kwon, Chen, Chow, Zhu, Wei, Egyed, Cotruta, Giang, Kirk, Rao, Black, Babar, Lo, Moreira, Martins, Sanseviero, Gonzalez, Gleicher, Warkentin, Mirrokni, Senter, Collins, Barral, Ghahramani, Hadsell, Matias, Sculley, Petrov, Fiedel, Shazeer, Vinyals, Dean, Hassabis, Kavukcuoglu, Farabet, Buchatskaya, Alayrac, Anil, Dmitry, Lepikhin, Borgeaud, Bachem, Joulin, Andreev, Hardin, Dadashi, and Hussenot]{gemma3}
Gemma Team, Aishwarya Kamath, Johan Ferret, Shreya Pathak, Nino Vieillard, Ramona Merhej, Sarah Perrin, Tatiana Matejovicova, Alexandre Ramé, Morgane Rivière, Louis Rouillard, Thomas Mesnard, Geoffrey Cideron, Jean bastien Grill, Sabela Ramos, Edouard Yvinec, Michelle Casbon, Etienne Pot, Ivo Penchev, Gaël Liu, Francesco Visin, Kathleen Kenealy, Lucas Beyer, Xiaohai Zhai, Anton Tsitsulin, Robert Busa-Fekete, Alex Feng, Noveen Sachdeva, Benjamin Coleman, Yi~Gao, Basil Mustafa, Iain Barr, Emilio Parisotto, David Tian, Matan Eyal, Colin Cherry, Jan-Thorsten Peter, Danila Sinopalnikov, Surya Bhupatiraju, Rishabh Agarwal, Mehran Kazemi, Dan Malkin, Ravin Kumar, David Vilar, Idan Brusilovsky, Jiaming Luo, Andreas Steiner, Abe Friesen, Abhanshu Sharma, Abheesht Sharma, Adi~Mayrav Gilady, Adrian Goedeckemeyer, Alaa Saade, Alex Feng, Alexander Kolesnikov, Alexei Bendebury, Alvin Abdagic, Amit Vadi, András György, André~Susano Pinto, Anil Das, Ankur Bapna, Antoine Miech, Antoine Yang, Antonia Paterson, Ashish
  Shenoy, Ayan Chakrabarti, Bilal Piot, Bo~Wu, Bobak Shahriari, Bryce Petrini, Charlie Chen, Charline~Le Lan, Christopher~A. Choquette-Choo, CJ~Carey, Cormac Brick, Daniel Deutsch, Danielle Eisenbud, Dee Cattle, Derek Cheng, Dimitris Paparas, Divyashree~Shivakumar Sreepathihalli, Doug Reid, Dustin Tran, Dustin Zelle, Eric Noland, Erwin Huizenga, Eugene Kharitonov, Frederick Liu, Gagik Amirkhanyan, Glenn Cameron, Hadi Hashemi, Hanna Klimczak-Plucińska, Harman Singh, Harsh Mehta, Harshal~Tushar Lehri, Hussein Hazimeh, Ian Ballantyne, Idan Szpektor, Ivan Nardini, Jean Pouget-Abadie, Jetha Chan, Joe Stanton, John Wieting, Jonathan Lai, Jordi Orbay, Joseph Fernandez, Josh Newlan, Ju~yeong Ji, Jyotinder Singh, Kat Black, Kathy Yu, Kevin Hui, Kiran Vodrahalli, Klaus Greff, Linhai Qiu, Marcella Valentine, Marina Coelho, Marvin Ritter, Matt Hoffman, Matthew Watson, Mayank Chaturvedi, Michael Moynihan, Min Ma, Nabila Babar, Natasha Noy, Nathan Byrd, Nick Roy, Nikola Momchev, Nilay Chauhan, Noveen Sachdeva, Oskar
  Bunyan, Pankil Botarda, Paul Caron, Paul~Kishan Rubenstein, Phil Culliton, Philipp Schmid, Pier~Giuseppe Sessa, Pingmei Xu, Piotr Stanczyk, Pouya Tafti, Rakesh Shivanna, Renjie Wu, Renke Pan, Reza Rokni, Rob Willoughby, Rohith Vallu, Ryan Mullins, Sammy Jerome, Sara Smoot, Sertan Girgin, Shariq Iqbal, Shashir Reddy, Shruti Sheth, Siim Põder, Sijal Bhatnagar, Sindhu~Raghuram Panyam, Sivan Eiger, Susan Zhang, Tianqi Liu, Trevor Yacovone, Tyler Liechty, Uday Kalra, Utku Evci, Vedant Misra, Vincent Roseberry, Vlad Feinberg, Vlad Kolesnikov, Woohyun Han, Woosuk Kwon, Xi~Chen, Yinlam Chow, Yuvein Zhu, Zichuan Wei, Zoltan Egyed, Victor Cotruta, Minh Giang, Phoebe Kirk, Anand Rao, Kat Black, Nabila Babar, Jessica Lo, Erica Moreira, Luiz~Gustavo Martins, Omar Sanseviero, Lucas Gonzalez, Zach Gleicher, Tris Warkentin, Vahab Mirrokni, Evan Senter, Eli Collins, Joelle Barral, Zoubin Ghahramani, Raia Hadsell, Yossi Matias, D.~Sculley, Slav Petrov, Noah Fiedel, Noam Shazeer, Oriol Vinyals, Jeff Dean, Demis Hassabis,
  Koray Kavukcuoglu, Clement Farabet, Elena Buchatskaya, Jean-Baptiste Alayrac, Rohan Anil, Dmitry, Lepikhin, Sebastian Borgeaud, Olivier Bachem, Armand Joulin, Alek Andreev, Cassidy Hardin, Robert Dadashi, and Léonard Hussenot.
\newblock Gemma 3 technical report, 2025.
\newblock URL \url{https://arxiv.org/abs/2503.19786}.

\bibitem[Tong et~al.(2024{\natexlab{a}})Tong, Brown, Wu, Woo, Middepogu, Akula, Yang, Yang, Iyer, Pan, Wang, Fergus, LeCun, and Xie]{camb}
Shengbang Tong, Ellis Brown, Penghao Wu, Sanghyun Woo, Manoj Middepogu, Sai~Charitha Akula, Jihan Yang, Shusheng Yang, Adithya Iyer, Xichen Pan, Ziteng Wang, Rob Fergus, Yann LeCun, and Saining Xie.
\newblock Cambrian-1: A fully open, vision-centric exploration of multimodal llms, 2024{\natexlab{a}}.
\newblock URL \url{https://arxiv.org/abs/2406.16860}.

\bibitem[Tong et~al.(2024{\natexlab{b}})Tong, Liu, Zhai, Ma, LeCun, and Xie]{eyes}
Shengbang Tong, Zhuang Liu, Yuexiang Zhai, Yi~Ma, Yann LeCun, and Saining Xie.
\newblock Eyes wide shut? exploring the visual shortcomings of multimodal llms.
\newblock \emph{2024 IEEE/CVF Conference on Computer Vision and Pattern Recognition (CVPR)}, pp.\  9568--9578, 2024{\natexlab{b}}.
\newblock URL \url{https://api.semanticscholar.org/CorpusID:266976992}.

\bibitem[Touvron et~al.(2022)Touvron, Cord, and Jégou]{deit}
Hugo Touvron, Matthieu Cord, and Hervé Jégou.
\newblock Deit iii: Revenge of the vit, 2022.
\newblock URL \url{https://arxiv.org/abs/2204.07118}.

\bibitem[Touvron et~al.(2023)Touvron, Lavril, Izacard, Martinet, Lachaux, Lacroix, Rozière, Goyal, Hambro, Azhar, Rodriguez, Joulin, Grave, and Lample]{llama}
Hugo Touvron, Thibaut Lavril, Gautier Izacard, Xavier Martinet, Marie-Anne Lachaux, Timothée Lacroix, Baptiste Rozière, Naman Goyal, Eric Hambro, Faisal Azhar, Aurelien Rodriguez, Armand Joulin, Edouard Grave, and Guillaume Lample.
\newblock Llama: Open and efficient foundation language models, 2023.
\newblock URL \url{https://arxiv.org/abs/2302.13971}.

\bibitem[Van~Horn et~al.(2015)Van~Horn, Branson, Farrell, Haber, Barry, Ipeirotis, Perona, and Belongie]{nabirds}
Grant Van~Horn, Steve Branson, Ryan Farrell, Scott Haber, Jessie Barry, Panos Ipeirotis, Pietro Perona, and Serge Belongie.
\newblock Building a bird recognition app and large scale dataset with citizen scientists: The fine print in fine-grained dataset collection.
\newblock In \emph{2015 IEEE Conference on Computer Vision and Pattern Recognition (CVPR)}, pp.\  595--604, 2015.
\newblock \doi{10.1109/CVPR.2015.7298658}.

\bibitem[Van~Horn et~al.(2018)Van~Horn, Mac~Aodha, Song, Cui, Sun, Shepard, Adam, Perona, and Belongie]{inat}
Grant Van~Horn, Oisin Mac~Aodha, Yang Song, Yin Cui, Chen Sun, Alex Shepard, Hartwig Adam, Pietro Perona, and Serge Belongie.
\newblock The inaturalist species classification and detection dataset.
\newblock In \emph{Proceedings of the IEEE conference on computer vision and pattern recognition}, pp.\  8769--8778, 2018.

\bibitem[Wah et~al.(2011)Wah, Branson, Welinder, Perona, and Belongie]{cub}
Catherine Wah, Steve Branson, Peter Welinder, Pietro Perona, and Serge Belongie.
\newblock The caltech-ucsd birds-200-2011 dataset.
\newblock Jul 2011.

\bibitem[Wu et~al.(2024{\natexlab{a}})Wu, Li, Xu, Yuan, Ding, Yang, Li, Zhang, Tong, Jiang, Ghanem, and Tao]{survey}
Jianzong Wu, Xiangtai Li, Shilin Xu, Haobo Yuan, Henghui Ding, Yibo Yang, Xia Li, Jiangning Zhang, Yunhai Tong, Xudong Jiang, Bernard Ghanem, and Dacheng Tao.
\newblock Towards open vocabulary learning: A survey, 2024{\natexlab{a}}.
\newblock URL \url{https://arxiv.org/abs/2306.15880}.

\bibitem[Wu et~al.(2024{\natexlab{b}})Wu, Li, Xu, Yuan, Ding, Yang, Li, Zhang, Tong, Jiang, Ghanem, and Tao]{wu2023open}
Jianzong Wu, Xiangtai Li, Shilin Xu, Haobo Yuan, Henghui Ding, Yibo Yang, Xia Li, Jiangning Zhang, Yunhai Tong, Xudong Jiang, Bernard Ghanem, and Dacheng Tao.
\newblock Towards open vocabulary learning: A survey.
\newblock \emph{T-PAMI}, 2024{\natexlab{b}}.

\bibitem[Wu et~al.(2025)Wu, Biamby, Quenum, Gupta, Gonzalez, Darrell, and Chan]{mirage}
Tsung-Han Wu, Giscard Biamby, Jerome Quenum, Ritwik Gupta, Joseph~E. Gonzalez, Trevor Darrell, and David Chan.
\newblock Visual haystacks: A vision-centric needle-in-a-haystack benchmark.
\newblock In \emph{The Thirteenth International Conference on Learning Representations}, 2025.
\newblock URL \url{https://openreview.net/forum?id=9JCNPFL1f9}.

\bibitem[Yang et~al.(2024)Yang, Yang, Hui, Zheng, Yu, Zhou, Li, Li, Liu, Huang, Dong, Wei, Lin, Tang, Wang, Yang, Tu, Zhang, Ma, Yang, Xu, Zhou, Bai, He, Lin, Dang, Lu, Chen, Yang, Li, Xue, Ni, Zhang, Wang, Peng, Men, Gao, Lin, Wang, Bai, Tan, Zhu, Li, Liu, Ge, Deng, Zhou, Ren, Zhang, Wei, Ren, Liu, Fan, Yao, Zhang, Wan, Chu, Liu, Cui, Zhang, Guo, and Fan]{qwen2}
An~Yang, Baosong Yang, Binyuan Hui, Bo~Zheng, Bowen Yu, Chang Zhou, Chengpeng Li, Chengyuan Li, Dayiheng Liu, Fei Huang, Guanting Dong, Haoran Wei, Huan Lin, Jialong Tang, Jialin Wang, Jian Yang, Jianhong Tu, Jianwei Zhang, Jianxin Ma, Jianxin Yang, Jin Xu, Jingren Zhou, Jinze Bai, Jinzheng He, Junyang Lin, Kai Dang, Keming Lu, Keqin Chen, Kexin Yang, Mei Li, Mingfeng Xue, Na~Ni, Pei Zhang, Peng Wang, Ru~Peng, Rui Men, Ruize Gao, Runji Lin, Shijie Wang, Shuai Bai, Sinan Tan, Tianhang Zhu, Tianhao Li, Tianyu Liu, Wenbin Ge, Xiaodong Deng, Xiaohuan Zhou, Xingzhang Ren, Xinyu Zhang, Xipin Wei, Xuancheng Ren, Xuejing Liu, Yang Fan, Yang Yao, Yichang Zhang, Yu~Wan, Yunfei Chu, Yuqiong Liu, Zeyu Cui, Zhenru Zhang, Zhifang Guo, and Zhihao Fan.
\newblock Qwen2 technical report, 2024.
\newblock URL \url{https://arxiv.org/abs/2407.10671}.

\bibitem[Yao et~al.(2024)Yao, Yu, Zhang, Wang, Cui, Zhu, Cai, Li, Zhao, He, et~al.]{minicpm}
Yuan Yao, Tianyu Yu, Ao~Zhang, Chongyi Wang, Junbo Cui, Hongji Zhu, Tianchi Cai, Haoyu Li, Weilin Zhao, Zhihui He, et~al.
\newblock Minicpm-v: A gpt-4v level mllm on your phone.
\newblock \emph{arXiv preprint arXiv:2408.01800}, 2024.

\bibitem[Yin et~al.(2024)Yin, Fu, Zhao, Li, Sun, Xu, and Chen]{Yin_2024}
Shukang Yin, Chaoyou Fu, Sirui Zhao, Ke~Li, Xing Sun, Tong Xu, and Enhong Chen.
\newblock A survey on multimodal large language models.
\newblock \emph{National Science Review}, 11\penalty0 (12), November 2024.
\newblock ISSN 2053-714X.
\newblock \doi{10.1093/nsr/nwae403}.
\newblock URL \url{http://dx.doi.org/10.1093/nsr/nwae403}.

\bibitem[Zhai et~al.(2023)Zhai, Mustafa, Kolesnikov, and Beyer]{siglip}
Xiaohua Zhai, Basil Mustafa, Alexander Kolesnikov, and Lucas Beyer.
\newblock Sigmoid loss for language image pre-training, 2023.
\newblock URL \url{https://arxiv.org/abs/2303.15343}.

\bibitem[Zhang et~al.(2024{\natexlab{a}})Zhang, Zhang, Dong, Zang, and Wang]{longclip}
Beichen Zhang, Pan Zhang, Xiaoyi Dong, Yuhang Zang, and Jiaqi Wang.
\newblock Long-clip: Unlocking the long-text capability of clip, 2024{\natexlab{a}}.
\newblock URL \url{https://arxiv.org/abs/2403.15378}.

\bibitem[Zhang et~al.(2024{\natexlab{b}})Zhang, Huang, Jin, and Lu]{vlmsurvey}
Jingyi Zhang, Jiaxing Huang, Sheng Jin, and Shijian Lu.
\newblock Vision-language models for vision tasks: A survey, 2024{\natexlab{b}}.
\newblock URL \url{https://arxiv.org/abs/2304.00685}.

\bibitem[Zhao et~al.(2017)Zhao, Puig, Zhou, Fidler, and Torralba]{ogovc}
Hang Zhao, Xavier Puig, Bolei Zhou, Sanja Fidler, and Antonio Torralba.
\newblock Open vocabulary scene parsing.
\newblock pp.\  2021--2029, 10 2017.
\newblock \doi{10.1109/ICCV.2017.221}.

\bibitem[Zhu et~al.(2025)Zhu, Wang, Chen, Liu, Ye, Gu, Tian, Duan, Su, Shao, Gao, Cui, Wang, Cao, Liu, Wei, Zhang, Wang, Xu, Li, Wang, Deng, Li, He, Jiang, Luo, Wang, He, Shi, Zhang, Shao, He, Xiong, Qu, Sun, Jiao, Lv, Wu, Zhang, Deng, Ge, Chen, Wang, Dou, Lu, Zhu, Lu, Lin, Qiao, Dai, and Wang]{internvl3}
Jinguo Zhu, Weiyun Wang, Zhe Chen, Zhaoyang Liu, Shenglong Ye, Lixin Gu, Hao Tian, Yuchen Duan, Weijie Su, Jie Shao, Zhangwei Gao, Erfei Cui, Xuehui Wang, Yue Cao, Yangzhou Liu, Xingguang Wei, Hongjie Zhang, Haomin Wang, Weiye Xu, Hao Li, Jiahao Wang, Nianchen Deng, Songze Li, Yinan He, Tan Jiang, Jiapeng Luo, Yi~Wang, Conghui He, Botian Shi, Xingcheng Zhang, Wenqi Shao, Junjun He, Yingtong Xiong, Wenwen Qu, Peng Sun, Penglong Jiao, Han Lv, Lijun Wu, Kaipeng Zhang, Huipeng Deng, Jiaye Ge, Kai Chen, Limin Wang, Min Dou, Lewei Lu, Xizhou Zhu, Tong Lu, Dahua Lin, Yu~Qiao, Jifeng Dai, and Wenhai Wang.
\newblock Internvl3: Exploring advanced training and test-time recipes for open-source multimodal models, 2025.
\newblock URL \url{https://arxiv.org/abs/2504.10479}.

\bibitem[Zhu et~al.(2024)Zhu, Gong, and Hoiem]{anytime}
Zhen Zhu, Yiming Gong, and Derek Hoiem.
\newblock Anytime continual learning for open vocabulary classification.
\newblock In \emph{Proceedings of the European Conference on Computer Vision (ECCV)}, 2024.

\end{thebibliography}
\bibliographystyle{iclr2025_conference}

\appendix
\newpage

\begin{wraptable}[8]{t}{0.5\textwidth} 
\vspace{-6mm}
\caption{Classification Results for remaining MLLMs}
\vspace{-3mm}
\centering
\scalebox{0.65}{
\footnotesize
\begin{tabular}{l|rrrrrlr}
& \rotatebox{90}{Birdsnap} & \rotatebox{90}{CUB} & \rotatebox{90}{iNat.}     & \rotatebox{90}{Ind. Birds}   & \rotatebox{90}{NABirds} & \rotatebox{90}{Average} \\
\midrule
InternVL-3~\cite{internvl3} & 12.7 & 14.9 & 4.0 & 0.5 & 14.4 & 9.3 \\
InternVL-3 + Direct-RAG & 27.0 & 27.0 & 12.6 & 25.8 & 28.3 & 24.1 \\
InternVL-3 + VR-RAG & 42.0 & 46.6 & 27.8 & 48.9 & 45.3 & 42.1 \\
\midrule
MiniCPM-V-2.6~\cite{minicpm} & 9.4 & 8.0 & 2.8 & 1.8 & 10.8 & 6.6 \\
MiniCPM-V-2.6 + Direct-RAG & 22.7 & 24.1 & 10.5 & 23.3 & 26.6 & 21.4 \\
MiniCPM-V-2.6 + VR-RAG & 36.0 & 40.8 & 24.6 & 44.7 & 40.8 & 37.4 \\
\midrule
Gemma-3n~\cite{qwen2} & 17.2 & 19.8 & 8.5 & 15.6 & 22.8 & 16.8 \\
Gemma-3n + Direct-RAG & 25.2 & 25.5 & 12.0 & 27.6 & 27.5 & 23.6 \\
Gemma-3n + VR-RAG & 44.8 & 52.7 & 32.2 & 61.1 & 48.3 & 47.8 \\
\midrule
\end{tabular}
}
\label{tab:supp_baselines}
\end{wraptable}

\section{Additional results with VR-RAG.}
\label{app:addvrrag}
In this section, we present results for the remaining MLLMs from~\cref{tab:supp_baselines} when integrated with VR-RAG, and compare them against direct-RAG, which uses CLIP as the retrieval module. As shown in~\cref{tab:supp_baselines}, VR-RAG consistently delivers clear performance gains across all five benchmarks.

\begin{wraptable}[15]{t}{0.5\textwidth} 
\vspace{-3mm}
\caption{Complete classification results for the FishNet and the \pok~dataset.}
\vspace{-3mm}
\centering
\scalebox{0.90}{
\footnotesize
\begin{tabular}{l|rrl}
& FishNet & \pok \\
\midrule
BioCLIP~\cite{bioclip} & 2.7 & 0.2 \\
CLIP~\cite{clip} & 2.8 & 60.5 \\
OpenCLIP~\cite{openclip} & 1.2 & 56.2 \\
SigLIP~\cite{siglip} & 3.1 & 75.5 \\
InternVL-3~\cite{internvl3} & 5.6 & 12.6 \\
MiniCPM-V-2.6~\cite{minicpm} & 1.3 & 17.1 \\
Qwen2VL~\cite{qwen2} & 3.9 & 30.1 \\
Gemma-3n~\cite{qwen2} & 4.7 & 30.8 \\
Qwen2.5VL~\cite{qwen25} & 5.0 & 29.4 \\
Qwen 2.5VL + direct RAG & 4.5 & 60.5 \\
\rowcolor{blond}
\midrule
Qwen 2VL + VR-RAG (ours) & 15.3 & 81.6 \\
\rowcolor{blond}
Qwen 2.5VL + VR-RAG (ours) & \textbf{18.7} & \textbf{86.3} \\
\midrule
\end{tabular}
}
\label{tab:supp_domain}
\end{wraptable}

\section{Complete Evaluation For FishNet and \pok.}
\label{app:fishpok}
In~\cref{tab:supp_domain}, we present the performance of all models on the FishNet and the \pok~datasets. All models struggle with the highly fine-grained and diverse FishNet dataset. Even in this challenging setting, the application of VR-RAG significantly improves the performance of MLLMs.

On the \pok~dataset, all Vision-Language Models except BioCLIP achieve superior performance compared to MLLMs. BioCLIP, having been exclusively trained on natural data from the Tree of Life~\cite{bioclip, treeoflife}, likely lacks exposure to non-natural domains, which explains its poor performance on this dataset. In contrast, the remaining VLMs perform exceptionally well, outperforming all MLLMs. However, when Qwen2.5-VL is augmented with VR-RAG, it achieves the best overall performance, demonstrating the effectiveness of our retrieval-augmented approach even in less-specialized domains compared to VLMs.

\begin{wraptable}[14]{t}{0.5\textwidth} 
\vspace{-4mm}
\caption{Retrieval results for the VLMs in VR-RAG when combining both text and visual representations, and two visual encoders. We report results for mRR@5}
\vspace{-3mm}
\centering
\scalebox{0.65}{
\footnotesize
\begin{tabular}{l|rrrrrlr}
& \rotatebox{90}{Birdsnap} & \rotatebox{90}{CUB} & \rotatebox{90}{iNat.}     & \rotatebox{90}{Ind. Birds}   & \rotatebox{90}{NABirds} & \rotatebox{90}{Average} \\
\midrule
CLIP$_t$ & 25.1 & 24.1 & 12.3 & 25.9 & 26.4 & 22.8 \\
CLIP$_v$ & 22.9 & 31.8 & 15.8 & 31.5 & 27.0 & 25.8 \\ 
CLIP$_{t+v}$ & 33.3 & 41.5 & 21.3 & 47.0 & 37.8 & 36.2\\
\midrule
OpenCLIP$_t$ & 24.8 & 23.9 & 13.7 & 16.8 & 23.9 & 20.6 \\
OpenCLIP$_v$ & 33.1 & 42.6 & 21.1 & 44.3 & 37.4 & 35.7\\
OpenCLIP$_{t+v}$ & 43.7 & 50.9 & 26.9 & 50.3 & 46.9 & 43.7\\
\midrule
SigLIP$_t$ & 27.9 & 29.6 & 17.2 & 30.4 & 29.6 & 26.9 \\
SigLIP$_v$ & 35.6 & 44.5 & 22.9 & 49.3 & 40.5 & 38.6 \\
SigLIP$_{t+v}$ & 43.2 & 49.9 & 27.0 & 53.9 & 47.4 & 44.3\\
\midrule
\end{tabular}
}

\label{tab:supp_vlm}
\end{wraptable}

\section{Evaluation with VLM using both encoders.}
\label{app:vlm_rerank_abl}

In~\cref{tab:supp_vlm}, we evaluate the performance of our VR-RAG encoders, examining how their multimodal representations, combining both text and visual encoders, enhance retrieval across all five benchmarks. The results demonstrate that this fusion of modalities consistently improves retrieval accuracy.

Furthermore, in~\cref{tab:rerankers}, we present a performance comparison when different visual encoders are used as re-rankers. We test the visual encoders from the VR-RAG pipeline, DinoV2~\cite{dinov2} and DeiT-III~\cite{deit}. The data clearly shows that DinoV2 significantly outperforms not only DeiT-III but also all the other encoders from various VLMs. This superior performance establishes DinoV2 as a strong candidate for re-ranking, capable of substantially improving the quality of initially retrieved chunks and refining the overall retrieval process.

\begin{table*}[t!]
\centering
\setlength\tabcolsep{10pt}
\caption{We evaluate different encoders as visual re-rankers by using them instead of DinoV2 in our pipeline.}

\scalebox{0.75}{
\begin{tabular}{cccccccc}
\toprule
 \multirow{2}{*}{\textbf{Re-ranker}}  & \textbf{Birdsnap} &\textbf{CUB} &\textbf{iNaturalist} &\textbf{Indian Birds} &\textbf{NABirds} &\textbf{Average} \\
 
\cmidrule(lr){2-7}

& mRR@1/5/10 & mRR@1/5/10 & mRR@1/5/10 & mRR@1/5/10 & mRR@1/5/10 & mRR@1/5/10 \\

\midrule
 
CLIP & 26.5/30.4/31.5 & 37.0/40.6/41.8 & 17.5/20.3/21.3 & 41.8/45.6/46.8 & 31.5/35.3/36.6 & 30.9/34.4/35.6\\
OpenCLIP & 32.7/35.9/36.9 & 41.2/45.0/46.0 & 19.3/22.0/23.0& 44.5/48.4/49.8& 35.9/39.5/40.7 & 34.7/38.2/39.3\\
SigLIP & 34.6/38.5/39.7 & 43.9/47.7/48.9& 21.3/24.4/25.4& 48.0/52.7/53.9& 39.3/43.5/44.7 & 37.4/41.4/42.5 \\
Deit-III & 18.5/21.5/23.1 & 22.5/25.5/27.0 & 11.8/14.2/15.5& 30.2/33.9/35.7& 21.1/24.1/25.8 & 20.8/23.8/25.4\\
DinoV2 & \textbf{48.9/52.6/53.8} & \textbf{58.0/62.3/63.1} & \textbf{34.8/38.7/39.8} & \textbf{68.1/72.9/73.6} & \textbf{52.0/56.6/57.6} & \textbf{52.3/56.6/57.6}\\

\bottomrule
\end{tabular}
}
\vspace{-0.1cm}
\label{tab:rerankers}
\end{table*}

\section{Ablation with retrieval components.}
\label{app:abl_mllm_ret}
In this section, we evaluate the contribution of each component in our retrieval pipeline within the VR-RAG framework using QWEN2.5-VL. Unlike~\cref{tab:retr_ablate} in the main paper, here we analyze the performance drop when specific components are removed from the complete pipeline. As shown in~\cref{tab:supp_ablate}, each additional component, OpenCLIP, SigLIP, multimodal fusion, and the re-ranker, contributes to consistent improvements across all five benchmarks. The complete VR-RAG pipeline achieves the highest average performance, demonstrating that all components are integral to its effectiveness.

\section{Why multi-modal retrieval when re-ranking with DinoV2?}
While the DinoV2 visual encoder exhibits strong unimodal performance(~\cref{tab:supp_vlm}), a critical question arises: can we bypass the multimodal retrieval step and rely solely on a cross-modal retrieval followed by DinoV2-based re-ranking? The effectiveness of this re-ranking hinges on the quality of the initial pool of top-ranked candidates. This approach should yield similar results to a full multimodal retrieval pipeline if the top-ranked candidates are already high-quality.

To investigate this, we evaluate VR-RAG without its multimodal representation, replacing the multimodal similarity score ($s^m$) with a cross-modal similarity ($s^c$). This is calculated by taking the dot product of the query image's visual features, $g(q)$, and the text chunk's features, $f(t)$, as shown in the equation below:
\begin{equation}
\begin{split}
    s^c = g(q)^\top f(t) \\
\end{split}
    \label{eqn:cr_sim}
\end{equation}
This initial step is followed by the re-ranking of the top r candidates in the same manner as the full pipeline(see~\cref{eqn:sims}). The results for all five avian datasets, the FishNet dataset, and the Pokémon dataset using Qwen 2.5VL are presented in Table \ref{tab:novis}.

\begin{wraptable}[7]{t}{0.6\textwidth} 
\vspace{-6mm}
\caption{Classification Results with cross-modal and multi-modal representation followed by DinoV2 reranker.}
\vspace{-3mm}
\centering
\scalebox{0.70}{
\footnotesize
\begin{tabular}{lrrrrrrrlr}
& \rotatebox{90}{Birdsnap} & \rotatebox{90}{CUB} & \rotatebox{90}{iNat}     & \rotatebox{90}{Ind Birds}   & \rotatebox{90}{NABirds} & \rotatebox{90}{FishNet} & \rotatebox{90}{\pok} & \rotatebox{90}{Average} \\
\midrule
VR-RAG (cross-modal) & 53.8 & 59.9 & 36.7 & 68.5 & 57.5 & 10.5 & 86.8 & 53.4\\
VR-RAG (multi-modal) & 51.9 & 60.3 & 35.5 & 72.2 & 56.3 & 18.7 & 86.3 & 54.5 \\
\midrule
\end{tabular}
}

\label{tab:novis}
\end{wraptable}

\begin{table*}[b!]
\centering
\setlength\tabcolsep{10pt}
\caption{Ablation study: Ablation study on retrieval components in VR-RAG with QWEN2.5-VL. We progressively add each component and report the accuracy across five benchmarks. Each module contributes to performance gains, with the full pipeline achieving the best results.}

\scalebox{0.6}{
\begin{tabular}{ccccccccccc}
\toprule
 \textbf{CLIP}  & \textbf{OpenCLIP} &
 \textbf{SigLIP} &
 \textbf{Multimodal} &
 \textbf{Re-Ranker} & \textbf{Birdsnap} &\textbf{CUB} &\textbf{iNat} &\textbf{Ind. Birds} &\textbf{NABirds} &\textbf{Average} \\
 
\midrule

{\color{green} \checkmark} & {\color{red} \ding{55}} & {\color{red} \ding{55}} & {\color{red} \ding{55}} & {\color{red} \ding{55}} & 29.5 & 29.2 & 13.2 & 29.5 & 31.7 & 26.6 \\

{\color{green} \checkmark} & {\color{green} \checkmark} & {\color{red} \ding{55}}& {\color{red} \ding{55}} & {\color{red} \ding{55}} & 38.9 & 44.3 & 18.9 & 36.2 & 43.5 & 36.4 \\

{\color{green} \checkmark} & {\color{green} \checkmark} & {\color{green} \checkmark} & {\color{red} \ding{55}} & {\color{red} \ding{55}} & 39.4 & 41.6 & 21.1 & 39.2 & 42.6 & 36.7 \\

{\color{green} \checkmark} & {\color{green} \checkmark} & {\color{green} \checkmark} & {\color{green} \checkmark} & {\color{red} \ding{55}} & 46.0 & 53.3 & 	26.9 & 58.7 & 51.3 & 47.2 \\

{\color{green} \checkmark} & {\color{green} \checkmark} & {\color{green} \checkmark} & {\color{green} \checkmark} & {\color{green} \checkmark} & {\textbf{51.9}} & {\textbf{60.3}} & {\textbf{35.5}} &
{\textbf{72.2}} &{\textbf{56.3}} &{\textbf{55.2}}\\

\bottomrule
\end{tabular}
}
\vspace{-0.1cm}
\label{tab:supp_ablate}
\end{table*}

The results show similar performance for both methods across all avian and Pokémon datasets. The average mRR@1 for these six datasets is 57.5\% with cross-modal retrieval and 57.8\% with multimodal retrieval. This similarity suggests that the initial cross-modal retrieval, in these cases, is robust enough to include high-quality candidates within the top 100 ranked species.

However, a significant performance gap emerges when the initial top-ranked candidates are of poor quality, as demonstrated by the FishNet dataset. In this case, the performance almost doubles in the multimodal setup. The initial cross-modal retrieval, which combines CLIP, OpenCLIP, and SigLip, yields a low mRR@1 of just 3.9\%. Re-ranking this poor set with DinoV2 improves performance to 8.2\% mRR@1.

In stark contrast, when we incorporate multimodal representations into the initial retrieval step, the mRR@1 score jumps to 14.5\%. With the subsequent DinoV2 re-ranking, this performance further increases to an impressive 17.8\% mRR@1, which is more than double the result of the unimodal approach. Furthermore, the inclusion of multimodal representation consistently improves results, even when using a single CLIP-like model, as detailed in Tables \ref{tab:supp_vlm} and \ref{tab:supp_ablate}.

\begin{wrapfigure}[20]{t}{0.5\textwidth} 
    \centering
    \vspace{-7mm}
    \includegraphics[width=\linewidth]{figs/rerank_ablation.png}
        \caption{Effect of re-ranking on retrieval performance. Average mRR at ranks 1, 5, and 10 is shown as a function of the number of candidates re-ranked. Performance improves substantially when re-ranking the first few dozen candidates, but the gains plateau quickly. Beyond 100 candidates, the improvement is marginal, motivating the choice of 100 as the default setting in the main experiments.}
    \label{fig:mrr_ab}
\end{wrapfigure}

\section{Re-ranking @ 100?}
~\Cref{fig:mrr_ab} demonstrates that re-ranking significantly improves retrieval performance, but the benefits diminish rapidly after a certain point. As the number of candidates considered for re-ranking increases from 0 to 100, the average mRR@1, mRR@5, and mRR@10 all show a steep and substantial increase. For example, the mRR@1 score increases from a baseline of around 44\% to over 51\% with just 50 candidates, indicating that the re-ranking process effectively moves the correct answer to a higher position. However, the curves for all three metrics begin to plateau after approximately 100 candidates, indicating that the performance gains become marginal. Beyond this threshold, the added computational cost of re-ranking hundreds of additional items does not yield a meaningful increase in accuracy. This analysis confirms the conclusion that 100 candidates represent an optimal trade-off between efficiency and performance.

\begin{wraptable}[18]{t}{0.5\textwidth} 
\vspace{-4mm}
\caption{Classification Results at Genus Level: We map all the predicted species by different models to their genus. This helps us to understand if the misclassification happens because the task is difficult or if the models are making unexpected mistakes. }
\centering
\scalebox{0.65}{
\footnotesize
\begin{tabular}{l|rrrrrlr}
& \rotatebox{90}{Birdsnap} & \rotatebox{90}{CUB} & \rotatebox{90}{iNat}     & \rotatebox{90}{Ind Birds}   & \rotatebox{90}{NABirds} & \rotatebox{90}{Average} \\
\midrule
BioCLIP~\cite{bioclip} & 43.4 & 46.2 & 33.6 & 29.2 & 21.5 & 32.0	 \\
CLIP~\cite{clip} & 37.3 & 40.5 & 44.1 & 23.6 & 18.6 & 32.8 \\
OpenCLIP~\cite{openclip} & 37.1 & 39.7 & 37.6 & 25.1 & 17.5 & 31.4 \\
SigLIP~\cite{siglip} & 40.6 & 38.6 & 44.4 & 28.7 & 27.7 & 36.0 \\
InternVL-3~\cite{internvl3} & 22.0 & 23.2 & 11.0 & 10.7 & 3.2 & 14.0 \\
MiniCPM-V-2.6~\cite{minicpm} & 13.9 &12.2 &6.3 &4.6 &16.3 & 10.7 \\
Qwen2VL~\cite{qwen2} & 45.8	& 46.8 & 23.1 & 29.9 & 51.9 & 39.5 \\
Gemma-3n~\cite{qwen2} & 28.5 & 30.2 & 16.7 & 21.1 & 32.2 & 25.7\\
Qwen2.5VL~\cite{qwen25} & 47.0 & 48.4 & 42.2 & 25.8 & 8.1 & 34.3 \\
\rowcolor{blond}
\midrule
Qwen 2VL + VR-RAG (ours) & 73.7 & 77.6 & 58.6 & 86.3 & 78.4 & 74.9 \\
\rowcolor{blond}
Qwen 2.5VL + VR-RAG (ours) & \textbf{74.8} & \textbf{79.9} & \textbf{60.3} & \textbf{86.6} & \textbf{79.2} & \textbf{76.2}\\
\midrule
\end{tabular}
}
\label{tab:supp_genus}
\end{wraptable}

\section{Analyzing the mis-classifications.}
In this section, we analyze the misclassifications to uncover interesting patterns. We leverage the taxonomic hierarchy by examining accuracy at the genus level. This approach reveals how frequently predictions fall within the same genus, as species within the same genus often share similar appearances. To calculate the genus accuracy, we map all the predicted species to the genus level and then calculate the accuracy. Instead of directly evaluating for genus, this approach allows us to analyze the behavior during species prediction. The results are presented in~\cref{tab:supp_genus}. Our VR-RAG framework model significantly enhances genus-level accuracy across all LMMs, indicating that the retrieved information frequently belongs to the same genus. Consequently, the task of species classification for the LMM still remains challenging, as species within the same genus share numerous attributes. This inherent ambiguity may explain why refining the summary improves species recognition. By providing more discriminative species descriptions, the model is better equipped to differentiate between species, even when they belong to the same genus.

As shown in~\cref{fig:3}, we present two examples where the top-5 retrieved candidates are incorrect. The retrieved candidates exhibit notable similarities to the query image, highlighting the challenging, fine-grained nature of retrieving the correct species from a large pool of 11,202 candidates.

In~\cref{fig:4}, we illustrate two instances where the correct species is successfully retrieved within the top 5 candidates. However, as is evident from the examples, the other four candidates also belong to the same genus. This high degree of similarity highlights the fine-grained nature of the retrieval task. The shared properties among these closely related species make it exceptionally challenging for the model to identify the correct candidate.
\\

\begin{wrapfigure}[16]{t}{0.5\textwidth} 
\vspace{-10mm}
    \centering
    \includegraphics[width=\linewidth]{figs/fig2_supp.pdf}
    \vspace{-5mm}
        \caption{Top-k species selection ablation. We report the results for Gemma-3n on all five benchmarks by varying the value of k from 5 to 15.}
    \label{fig:2_supp}
    \vspace{-5mm}
\end{wrapfigure} 

\section{Top-k selection.}
\label{app:topk}
In~\cref{fig:2}, we showed the impact of increasing the number of top-k species candidates from $k = 5$ to $k=15$ on the performance of the Qwen2.5-VL model. In~\cref{fig:2_supp}, we show the performance for the Gemma-3n model, and it shows a similar pattern as well. The average performance decreases from 47.8\% at $k=5$ to 47.0\% at $k=10$, and then to 46.3\% at $k=15$. And similar to Qwen2.5-VL, Gemma-3n also performs better at $k=15$ with 47.0\% rather than feeding in 5 Wikipedia articles of top-5 candidates, which results in an average of 33.6\% accuracy. The drop is more severe for Gemma-3n as compared to Qwen2.5-Vl, highlighting the weakness of Gemma-3n struggling with bigger contexts.


\section{The Re-ranking module}

\begin{wrapfigure}[7]{t}{0.5\textwidth} 
\vspace{-15mm}
    \centering
    \includegraphics[width=\linewidth]{figs/rerank.pdf}
        \caption{The visual representation of the re-ranker module.}
    \label{fig:rerank}
    \vspace{-5mm}
\end{wrapfigure}

The re-ranking module begins by selecting the top-$r$ candidate species. For each of these species, the intra-modal similarity is calculated between the anchor images and the query image according to~\cref{eqn:sims}. A visual illustration of this process is provided in~\cref{fig:rerank}.

\section{Human evaluation}
\label{app:human}

\subsection{Matching Description With Images}

In this section, we analyze the discriminatory power of our generated species descriptions through a qualitative human study. Annotators of the forced-choice matching test also classified each of the 200 tasks as easy, medium, or hard based on the perceived time and effort required.

As illustrated in~\cref{fig:easy}, tasks classified as easy typically featured a description with a visual trait unique to the correct species, making the correct image immediately stand out. Tasks classified as medium, shown in~\cref{fig:medium}, required more focused attention, as the key distinguishing features were more subtle. For tasks classified as hard, shown in~\cref{fig:hard}, the distractors were extremely similar to the correct image, often differing by a single, minute attribute that required users to zoom in to discern the difference.

In~\cref{fig:ambig}, we look at the cases where the annotators are inconclusive about their final decision, and in~\cref{fig:wrong}, where annotators made incorrect choices. 
These scenarios occur under similar conditions, such as when a key visual trait was obscured in the image or when an annotator overlooked a critical detail in the description. For instance, in the first row of~\cref{fig:ambig}, the annotator states that a clearer image of one of the confusing options would greatly solve the issue. And, in the second row of~\cref{fig:wrong}, the annotator incorrectly matches a bird because they mistook the "pale crown stripe" as the white line above the bird's eyes, forgetting that the pale strip is different as evident from the next sentence in description that tells us the pale stripe complements the white line and this is a feature present only in the correct image.

Ultimately, the difficulty of this fine-grained matching task shows the high quality and discriminatory power of our generated descriptions. They consistently include the specific, and often subtle, attributes necessary to differentiate species even in visually challenging scenarios.
 
\begin{figure}[h]
    \centering
    \includegraphics[width=\linewidth]{figs/easy.pdf}
        \caption{Examples of description matching where the annotators classified the cases as easy. Wrong options are highlighted in red, and the correct option is highlighted in green.}
    \label{fig:easy}
\end{figure}  
\clearpage 

\begin{figure}[h]
    \centering
    \includegraphics[width=\linewidth]{figs/medium.pdf}
        \caption{Examples of description matching where the annotators classified the cases as medium. Wrong options are highlighted in red, and the correct option is highlighted in green.}
    \label{fig:medium}
\end{figure}  
\clearpage

\begin{figure}[h]
    \centering
    \includegraphics[width=\linewidth]{figs/hard.pdf}
        \caption{Examples of description matching where the annotators classified the cases as hard. Wrong options are highlighted in red, and the correct option is highlighted in green.}
    \label{fig:hard}
\end{figure}
\clearpage 

\begin{figure}[h]
    \centering
    \includegraphics[width=\linewidth]{figs/ambig.pdf}
        \caption{Examples of description matching where the annotators are unable to pick one single option. Wrong options are highlighted in red, and the correct option is highlighted in green.}
    \label{fig:ambig}
\end{figure}
\clearpage 

\begin{figure}[h]
    \centering
    \includegraphics[width=\linewidth]{figs/wrong.pdf}
        \caption{Examples of description matching where the annotators pick the wrong option.}
    \label{fig:wrong}
\end{figure}
\clearpage 

\subsection{Reliability Methodology}
The primary goal of our inter-annotator reliability analysis was to quantify the consistency of human judgments on the quality of the generated descriptions. To provide a direct and interpretable measure of consensus among the annotators, we use Quadratic Weighted Agreement (QWA), which is a direct application of the weighting principle from~\cite{Cohen1968WeightedKN}, simplified by not correcting for chance agreement. The approach is described in~\cref{alg:qwa}.



\begin{algorithm}
\caption{Calculation of Quadratic Weighted Agreement (QWA)}
\label{alg:qwa}
\begin{algorithmic}[1]
\State \textbf{Input:} A set of items $I$, where each item $i \in I$ has a list of scores $R_i = [s_1, s_2, ..., s_n]$ from $n$ annotators. The scoring scale has a minimum value $S_{min}$ and a maximum value $S_{max}$.
\State \textbf{Output:} The overall Quadratic Weighted Agreement score, $QWA_{overall}$.

\vspace{1em} 

\State Initialize an empty list, $A_{items}$, to store the agreement score for each item.
\State $D_{max} \gets (S_{max} - S_{min})^2$ \Comment{Calculate max squared difference once}

\ForAll{item $i$ in $I$}
    \State Let $R_i$ be the list of scores for item $i$.
    \State Initialize an empty list, $W_{pairs}$, to store weights for the current item.
    \State Let $P_i$ be the set of all unique pairs of scores from $R_i$.
    
    \ForAll{pair $(s_a, s_b)$ in $P_i$}
        \State $D_{pair} \gets (s_a - s_b)^2$
        \State $W_{pair} \gets 1 - \frac{D_{pair}}{D_{max}}$ \Comment{Quadratic weighting formula}
        \State Add $W_{pair}$ to the list $W_{pairs}$.
    \EndFor
    
    \If{$W_{pairs}$ is not empty}
        \State $A_{item} \gets \text{average}(W_{pairs})$ \Comment{Mean agreement for the item}
        \State Add $A_{item}$ to the list $A_{items}$.
    \EndIf
\EndFor

\State $QWA_{overall} \gets \text{average}(A_{items})$ \Comment{Grand mean across all items}
\State \Return $QWA_{overall}$
\end{algorithmic}
\end{algorithm}

\section{Qualitative Description Samples}
We show qualitative samples with images of species and their descriptions from our data. We show samples for various birds in~\cref{fig:6}, and~\cref{fig:7}. For our curated \pok~dataset, we show examples in~\cref{fig:8}.

\begin{figure*}[h]
    \centering
    \includegraphics[width=\linewidth]{figs/ret_1.pdf}
        \caption{Examples of top-5 retrieved candidates that do not contain the correct species description. The retrieved images and their corresponding text descriptions show significant visual and textual similarities to the query image, highlighting the fine-grained and challenging nature of the retrieval task.}
    \label{fig:3}
\end{figure*}

\begin{figure*}[h]
    \centering
    \includegraphics[width=\linewidth]{figs/ret_2.pdf}
        \caption{Examples of top-5 retrieved candidates that contain the correct species description ranked. The other four candidates belong to the same genus as the query image and exhibit significant visual and textual similarities to it, highlighting the fine-grained and challenging nature of the retrieval task.}
    \label{fig:4}
\end{figure*}

\begin{figure*}[h]
    \centering
    \includegraphics[width=\linewidth]{figs/q_desc1.pdf}
        \caption{Examples of bird species with their descriptions.}
    \label{fig:6}
\end{figure*}  

\begin{figure*}[h]
    \centering
    \includegraphics[width=\linewidth]{figs/q_desc2.pdf}
        \caption{Examples of more bird species with their descriptions.}
    \label{fig:7}
\end{figure*}

\begin{figure*}[h]
    \centering
    \includegraphics[width=\linewidth]{figs/pok_1.pdf}
        \caption{\pok~species with their descriptions.}
    \label{fig:8}
\end{figure*}

\end{document}